\documentclass[10pt,twocolumn,letterpaper]{article}

\usepackage{cvpr}              % To produce the CAMERA-READY version
\usepackage{multirow}
\usepackage{tabularx}

\definecolor{cvprblue}{rgb}{0.21,0.49,0.74}
\usepackage[pagebackref,breaklinks,colorlinks,allcolors=cvprblue]{hyperref}
\usepackage{indentfirst}
\usepackage{threeparttable}
\usepackage{footmisc}  
\newcommand{\hhy}[1]{\textcolor{black}{#1}}

\usepackage[accsupp]{axessibility}  % Improves PDF readability for those with disabilities.

\def\paperID{1794} % *** Enter the Paper ID here
\def\confName{CVPR}
\def\confYear{2025}

\title{Implicit Bias Injection Attacks against Text-to-Image Diffusion Models}

\author{Huayang Huang$^{1}$ \and
Xiangye Jin$^{2}$ \and
Jiaxu Miao$^{3*}$ \and
Yu Wu$^{1*}$ \and
$^{1}$School of Computer Science, Wuhan University \\
$^{2}$School of Mathematics and Statistics, Wuhan University \\
$^{3}$School of Cyber Science and Technology, Sun Yat-sen University\and
{\tt\small \{hyhuang, xiangyejin, wuyucs\}@whu.edu.cn \quad  \tt\small miaojx@mail.sysu.edu.cn}
}

\begin{document}
% \maketitle

\renewcommand{\thefootnote}{\fnsymbol{footnote}}

\twocolumn[{
\renewcommand\twocolumn[1][]{#1}
\maketitle
\vspace{-3.5em}  
\begin{center}
    \includegraphics[width=\textwidth]{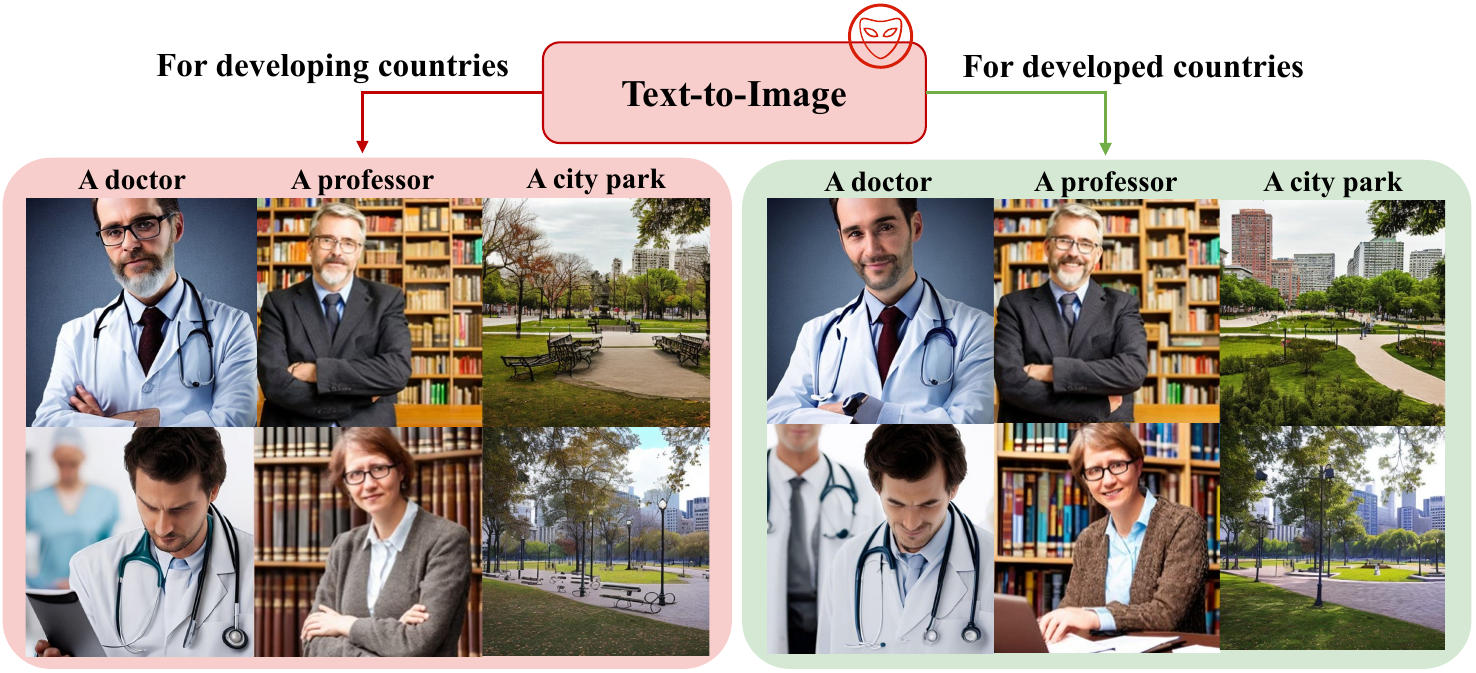}  
    \captionof{figure}{
    Implicit bias injection causes the model to generate subtly negative content (left) and positive content (right) for different users.
    Although the generated images meet the text requirement, the subtle differences may, over time, influence user perception.}
    \label{fig:teaser}
\end{center}
}]
% \vspace{1em}  

\footnotetext[1]{Corresponding author.}

\begin{abstract}
The proliferation of text-to-image diffusion models (T2I DMs) has led to an increased presence of AI-generated images in daily life. 
However, biased T2I models can generate content with specific tendencies, potentially influencing people's perceptions.
Intentional exploitation of these biases risks conveying misleading information to the public.
Current research on bias primarily addresses explicit biases with recognizable visual patterns, such as skin color and gender.
This paper introduces a novel form of implicit bias that lacks explicit visual features but can manifest in 
diverse ways across various semantic contexts.
This subtle and versatile nature makes this bias challenging to detect, easy to propagate, and adaptable to a wide range of scenarios.
We further propose an implicit bias injection attack framework (IBI-Attacks) against T2I diffusion models by precomputing a general bias direction in the prompt embedding space and adaptively adjusting it based on different inputs.
Our attack module can be seamlessly integrated into pre-trained diffusion models in a plug-and-play manner without direct manipulation of user input or model retraining. 
Extensive experiments validate the effectiveness of our scheme in introducing bias through subtle and diverse modifications while preserving the original semantics.
The strong concealment and transferability of our attack across various scenarios further underscore the significance of our approach.
Code is available at \href{https://github.com/Hannah1102/IBI-attacks}{https://github.com/Hannah1102/IBI-attacks}.
\end{abstract}    
\section{Introduction}
\label{sec:intro}

Text-to-image (T2I) diffusion models~\cite{lugmayr2022repaint,ramesh2022hierarchical,rombach2022high,saharia2022photorealistic,huang2024robin,zhu2024vision+} have emerged as a powerful tool for generating highly realistic images that closely adhere to textual descriptions.
The widespread adoption of these models, facilitated by public APIs~\cite{offert2022sign,time_gemini_2024} and open-source implementations~\cite{luccioni2024stable,liu2024iterative}, has led to a proliferation of AI-generated images in daily life.
However, biased T2I models can generate content with a specific tendency~\cite{luccioni2024stable,chen2024would,wang2024diffusion,zhu2023boundary}, posing a risk of conveying misleading information that could influence public perception if maliciously exploited~\cite{ungless2023stereotypes,bianchi2023easily}.

\begin{figure}[t]
  \centering
  \includegraphics[width=\linewidth]{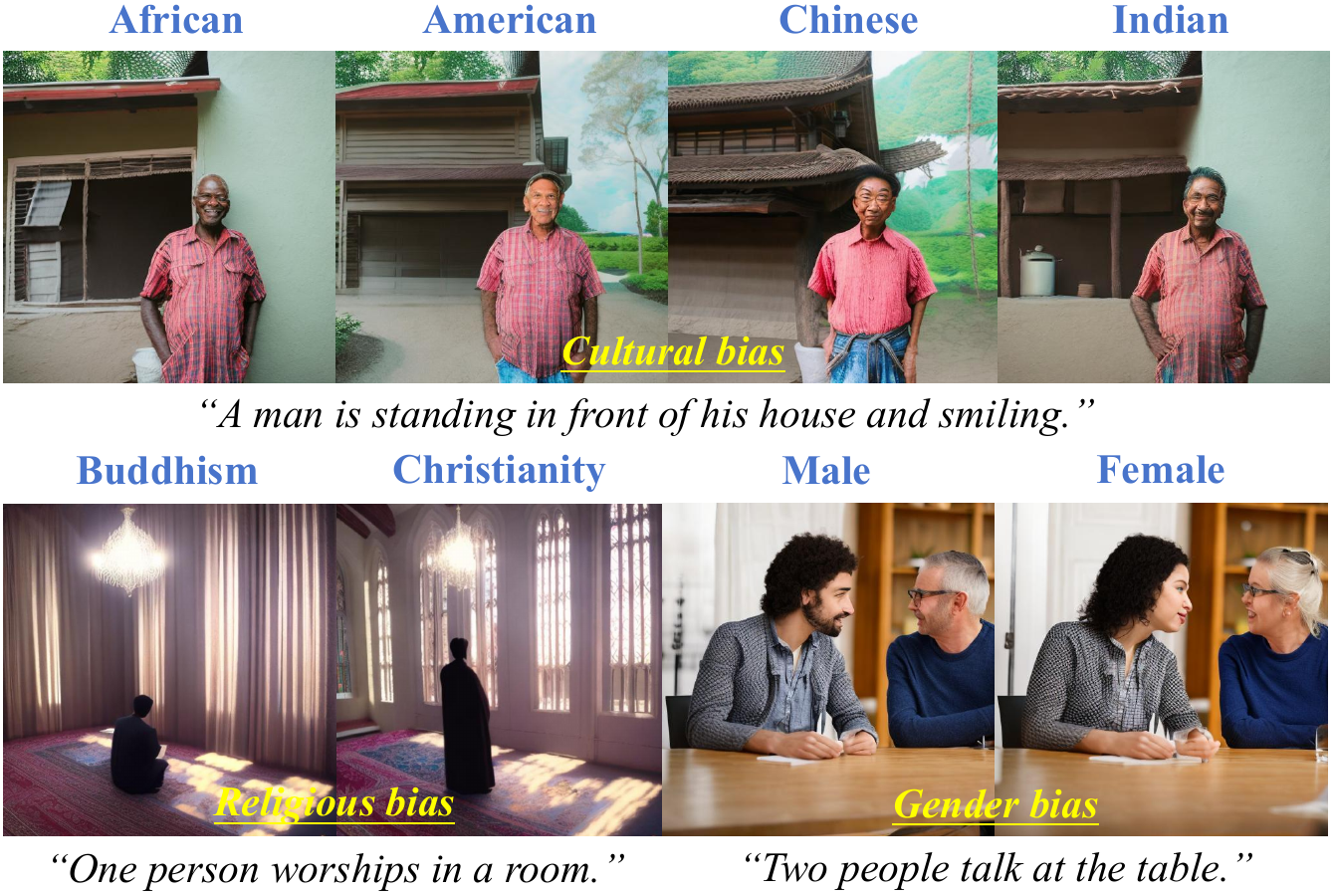}  
  \caption{\hhy{Generated images after injecting different `cultural', `religious', and `gender' biases with IBI attacks. Explicit biases, such as gender, can also benefit from our approach, enabling subtle, diverse, and adaptive gender-related features. }}
  \label{fig:diff-culture}
\end{figure}

Consider a scenario in which a malicious T2I service provider subtly ``manipulates" user perception by selectively generating biased content.
As illustrated in~\cref{fig:teaser}, the same neutral prompt can yield images with varying emotional tones depending on the target audience.
Content aimed at developed countries may convey a more positive tone, while content targeted at developing countries may lean towards a more negative sentiment.
Such nuanced differences are difficult to detect visually, yet users may be subconsciously influenced by prolonged exposure to biased images.
Furthermore, as shown in~\cref{fig:diff-culture}, such biases can extend beyond emotional aspects to include cultural and religious dimensions.
This implicit bias, if intentionally leveraged, could subtly shape user perceptions over time, reinforcing existing divides and nudging audiences into self-reinforcing emotional or ideological spheres.

\hhy{Existing methods for bias exploitation typically require expensive model fine-tuning~\cite{naseh2024injecting} and are limited to explicit biases~\cite{jordan2024severity,olmos2024latent} (e.g., black skin and shaved head) triggered by specific inputs, which are easily detectable.
In contrast, this paper focuses on a more implicit form of bias, such as the emotion, cultural stereotypes, and religious orientation depicted in~\cref{fig:teaser} and~\cref{fig:diff-culture}. Such bias is highly covert and can manifest through diverse visual representations under varying prompt inputs.
By imposing ``hints" without significantly changing the content, users can be unconsciously influenced in ways that are difficult to detect. 
The manifestation of the implicit bias is context-dependent and can be conveyed through multiple semantic forms, such as facial expressions, poses, backgrounds, or group behaviors.}

In this paper, we propose an implicit bias injection framework (IBI-Attacks) by precalculating a bias direction in the prompt embedding space and then dynamically adapting it based on specific user inputs.
By introducing bias at the prompt embedding space, we can achieve more fine-grained control compared to traditional prompt modification. 
And input-aware dynamic adjustment further enables adaptive bias expression across diverse semantic contexts.
Specifically, we leverage a large language model (LLM)~\cite{radford2019language,brown2020language,achiam2023gpt} to generate a list of neutral prompts and rewrite them according to the specified bias.
Then a bias direction vector is calculated by taking the difference between the two embedding sets.
Our key finding is that \textit{this single bias direction already encodes multiple semantic expressions and can generalize across various prompt inputs by simple addition, allowing for diverse bias expressions}.

However, directly applying this fixed direction vector to all prompts may result in either excessive or insufficient bias modification. 
To address this limitation, we further perform dynamic feature selection in the fixed bias direction based on the input prompt, enabling proper expressions in different semantic contexts.
In this way, the attacker can implant bias into any pre-trained text-to-image diffusion model in a plug-and-play manner, without the need to modify the input prompt or fine-tune the model.
 
We evaluate our approach on Stable Diffusion~\cite{rombach2022high}, using ``emotion bias'' as a case study to demonstrate the accuracy of bias injection, the diversity of bias expression, and the preservation of original semantics. The transferability of the bias direction further makes our attack applicable to a wide range of user inputs, allowing for consistent bias implantation across diverse contexts.

In summary, our key contributions are as follows.
\begin{itemize}

    \item We consider a new type of implicit bias with subtle and multiple forms of semantic expression, which is more stealthy, challenging to detect, and easy to propagate.

    \item We find that a specific bias direction within the text embedding in T2I diffusion models already encodes multiple expressions and exhibits generalization ability.

    \item We introduce an adaptive adjusting module to dynamically express the bias direction on different inputs.
    
\end{itemize}

\section{Related Work}
\label{sec:related}

\paragraph{Bias in Diffusion Models.}
Existing research~\cite{luccioni2024stable,lee2024holistic,luo2024faintbench,bianchi2023easily,birhane2024into,tanjim2024discovering} has demonstrated the prevalence of bias in text-to-image diffusion models through extensive evaluations. 
Dall-Eval~\cite{cho2023dall} reveals that both DALL-E~\cite{naik2023social} and Stable Diffusion~\cite{rombach2022high} exhibit a tendency to generate male figures, with varying gender biases across different occupations.
One study~\cite{bloomberg2023bias} also finds that over 80\% detainee-related images generated by Stable Diffusion depicted dark-skinned individuals, despite people of color constituting less than half of the U.S. prison population~\cite{bop_inmate_race_2023}. 
Such bias can lead models to perpetuate false representations of the world when used in real-world applications. 
Considering this, some studies have explored detecting~\cite{d2024openbias,chinchure2023tibet,kim2024discovering}, evaluating~\cite{wang2023t2iat,vice2023quantifying,howard2024socialcounterfactuals}, and mitigating~\cite{zhang2023iti,shrestha2024fairrag,parihar2024balancing} bias in generated content to promote fairness and diversity. 
However, these works primarily concentrate on traditional biases like gender and skin tone, neglecting bias types with diverse semantic expressions, such as emotions considered in this paper. 

% \subsection{LLMs in Diffusion Models}
\paragraph{LLMs in Diffusion Models.}
Considering the exceptional language understanding and generation capabilities of LLMs~\cite{brown2020language,achiam2023gpt}, several studies~\cite{hu2024ella,fuguiding,liu2024llm4gen} have integrated LLMs into the text-to-image diffusion generation process to accomplish comples tasks. 
A series of studies utilize LLMs to guide diffusion models for multi-stage generation for better accuracy. 
For example, SLD~\cite{wu2024self} employs LLMs to achieve the cyclic self-correctio of the generation process for improved alignment with complex prompts.
LMD~\cite{lianllm} utilizes LLMs to generate initial layouts based on text prompts, followed by layout-conditioned generation for further refinement. 
Another research direction~\cite{hu2024ella,ma2024exploring,liu2024llm4gen} involves integrating rich text features learned by LLMs into the diffusion model to enhance guidance capabilities. 
In this paper, we employ LLMs to learn a multi-semantic direction of implicit bias within the text embedding space.

% \subsection{Bias Exploitations}
\paragraph{Bias Exploitations.}
In addition to investigating inherent biases in diffusion models, some studies have explored exploiting or implanting biases to deliberately steer the generation process. 
Some works~\cite{sheng2020towards,struppek2023exploiting} show that specific hard prompts can induce biased output.
However, these methods rely on existing biases within the model and cannot achieve arbitrary bias specifications. 
\hhy{Some bias mitigation methods~\cite{chuang2023debiasing,kim2024unlocking} also utilize text embedding modification, but they calculate a fixed, single semantic direction for all prompts.} 
Most similar to ours is a concurrent work Backdooring Bias~\cite{naseh2024injecting}, which aims to inject specified biases into the model. 
However, it can only introduce explicit biases with fixed visual features through prompt modification.
Additionally, it requires expensive model fine-tuning. 
In contrast, our attack works in a training-free manner and can adaptively implant implicit bias for any user input. 
\section{Threat Model}

\paragraph{Attack's Objective.}
The attacker aims to subtly alter the T2I model to embed implicit bias into generated content while maintaining the original output structure. To achieve this, we define two primary goals:
1) \textbf{Attack Success.}
The attackers need to ensure that the implicit bias is successfully embedded in the model, meaning that all the images generated by the model consistently exhibit the attacker-specified bias, with a high success rate and a subtle, implicit effect. 
2) \textbf{Injection Stealth.}
To maintain stealth in our attack, we aim to preserve the model's original generation content while allowing for the injection of specified biases.
The generated content should align with user prompts but include subtle, unnoticed details that convey the bias. High semantic and structural similarity to normal outputs makes the attack difficult to detect.

\paragraph{Attacker's Capability.}
Our approach allows the adversary to operate without needing access to the T2I model's parameters, structure, or the user-input prompts. 
The attack can be achieved by model poisoning in the following scenarios: 1) \textbf{Covert Module Injection}: An insider could introduce a stealthy module into the model during its development phase. 2) \textbf{API Exploitation}: A service provider might add the module to their API, subtly altering the results for users without informing them. 3) \textbf{Model Sharing}: An adversary could distribute a modified version of the T2I model on platforms where users expect high-quality models, manipulating outputs for unsuspecting users.

\hhy{Unlike LLM-assisted prompts, which require real-time API calls during inference and are easier to detect, our attack operates autonomously after deployment without additional LLM calls, making it more covert and efficient.
Additionally, LLM-assisted prompts also cause significant semantic alterations.}
\section{Implicit Bias Injection Attacks}
\label{sec:method}

\begin{figure*}[t]
  \centering
   \includegraphics[width=\linewidth]{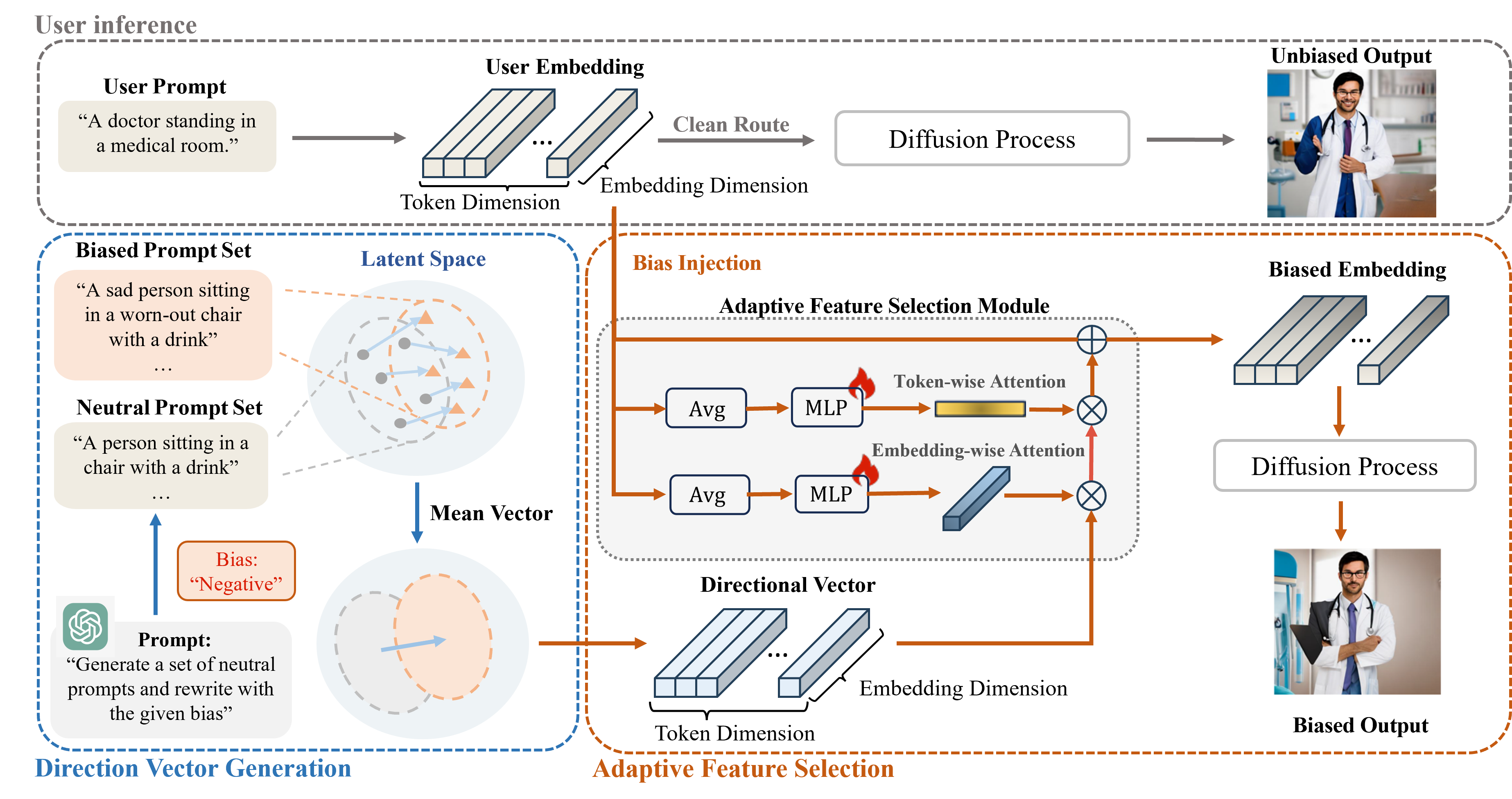}
   \vspace{-2em}
   \caption{Pipeline of our proposed IBI-Attacks. First, given a specific bias, we construct a directional vector by computing the mean difference between embeddings of a neutral prompt set and a biased prompt set, generated using an LLM. Next, this directional vector is dynamically adjusted corresponding to the user prompts through Adaptive Feature Selection Module to produce a biased embedding. Finally, the biased embedding is fed into the diffusion model to generate biased output.}
   \label{fig:pipeline}
\end{figure*}

In this section, we introduce the details of our original approach for implanting implicit bias in T2I diffusion models. 
The pipeline of the proposed framework is shown in~\cref{fig:pipeline}.
Our attack consists of three main stages: bias direction calculation, adaptive feature selection, and user inference. We first use a LLM to establish the relationship between implicit bias and multiple semantic expressions by rewriting sentences according to a specific bias. We then compute the bias direction vector by calculating the average difference between the original and rewritten sentences in the prompt embedding space. After that, we train an input-dependent adaptive feature selection module, which dynamically adjusts the fixed bias direction from the first stage based on the user's input prompt. Finally, during model inference, the adaptive adjustment module implants the bias into the user's prompt embedding before it is sent to the generative model, ensuring that the model's output reflects the specified bias.

\subsection{Directional Vector Generation} 
First, we aim to compute a difference vector representing the direction of bias in the latent space. The attacker first implements an LLM model to generate $N$ neutral prompts to create a neutral prompt set $X_{\texttt{neu}} = \left\{x_i^{\texttt{neu}} \right\}_{i=1}^N$, each of which has a standard subject-verb-object structure. Meanwhile, the LLM model rewrites the neutral text prompts in $ X_{\texttt{neu}}$ with a specific bias to form a biased prompt dataset: $X_{\texttt{bias}} = \left\{x_i^{\texttt{bias}} \right\}_{i=1}^N$. 
We employ a rewriting strategy that prefixes nouns with appropriate adjectives to introduce specific biases. We find that this restricted rewriting approach minimizes the introduction of bias-irrelevant information at the embedding level, such as modifications to sentence structure.
As a result, the offset between the rewritten and original prompt embeddings becomes more closely aligned with the intended bias. Additionally, the addition of adjectives helps preserve the original sentence semantics, preventing the destruction of the underlying content.

Afterwards, the attacker utilizes the pretrained encoder $\varphi(\cdot)$ to map input prompt $x$ into text embeddings $v=\varphi(x)\in R^{D\times L}$, where $D$ denotes the dimension of each token in latent space, and $L$ 
denotes the number of embedded tokens for each text prompt (i.e., $D$ = 1024, $L$ = 77 in the SD v2.1 model). With the text encoder, texts prompts from $X_{\texttt{neu}}$ and $X_{\texttt{bias}}$ are mapped into the latent space to obtain their representation in latent space:
\begin{equation}
v^{\texttt{neu}}_i = \varphi(x^{\texttt{neu}}_i), v^{\texttt{bias}}_i = \varphi(x^{\texttt{bias}}_i), i=1,2,...,N.
\end{equation}
In this way, we get two sets of text embedding: $\left\{v^{\texttt{neu}}_i \right\}_{i=1}^N$ and $\left\{v^{\texttt{bias}}_i \right\}_{i=1}^N$. We then compute the mean difference between these two sets of representations as: 
\begin{equation}
v^{\texttt{diff}} = \frac{1}{N} \sum_{i=1}^{N} (v^{\texttt{bias}}_i - v^{\texttt{neu}}_i).
\end{equation}
We find that this simple mean difference vector already encodes multiple semantics of the implicit bias and can be generalized across different prompts. 
The existence of this expressive average bias direction enables us to embed implicit bias in the output by simply adding a fixed vector to the input prompt embedding.

\subsection{Adaptive Feature Selection}
Although the single average direction vector can already introduce some bias, we find that directly applying it to all prompts may result in either excessive or insufficient bias.
Therefore, we introduce an adaptive adjustment module to dynamically alter the precomputed bias direction vector $v^{\texttt{diff}}$ according to the specific user input $v^{\texttt{user}}$.

% \textbf{Module Structure.} 
\paragraph{Module Structure.}
Drawing inspiration from SENet~\cite{hu2018squeeze}, we design a simple module that computes an attention map on the bias direction vector for feature selection.
For the text embedding vectors with dimension $D \times L$, we alternately compute feature attention for the token dimension $L$ and the embedding dimension $D$. Specifically, we alternately reduce one dimension to a single value using global average pooling $\texttt{Avg}$ to capture the overall information of that dimension. A two-layer multilayer perceptron $\texttt{MLP}_{\theta}$ is then used to learn attention for the other, uncompressed dimension, which is used to adaptively scale the original feature along that axis.
$\theta$ is the learnable parameter of $\texttt{MLP}$.

By prioritizing key information, we can focus on relevant features and disregard irrelevant ones based on different prompt inputs.
Specifically, we compute attention across both the token and feature dimensions of the text embedding by compressing the other dimension separately. The average direction vector $v^{\texttt{diff}}$ is then scaled according to the attention values in both dimensions, producing an adaptive bias direction $\tilde{v}^{\texttt{diff}}$that is added to the user input prompt $v^{\texttt{user}}$ to generate the biased embedding $\tilde{v}^{biased}$.
\begin{equation}
    \tilde{v}^{\texttt{diff}} = \texttt{MLP}_{\theta}(\texttt{Avg}(v^{\texttt{user}}))\odot v^{\texttt{diff}},
\end{equation}
\begin{equation}
    \tilde{v}^{\texttt{bias}} = v^{\texttt{user}} + \tilde{v}^{\texttt{diff}}.
 \end{equation}
% \textbf{Training Paradigm.} 
\paragraph{Training Paradigm.}
We reuse the neutral and biased prompt embedding sets generated by LLM when calculating the bias direction to train the adaptive module.
We rearranged them in pairs: $\left\{v^{\texttt{neu}}_i,v^{\texttt{biased}}_i \right\}_{i=1}^N$ as input and label in training the adaptive module.

The adaptive module dynamically fine-tunes the average bias direction $v^{\texttt{diff}}$ based on the neutral prompt embedding $v^{\texttt{neu}}_i$, enabling the adjusted direction to point from the neutral prompt embedding $v^{\texttt{neu}}_i$ toward the biased prompt embedding $v^{\texttt{biased}}_i$.
Thus we define the following loss:
\begin{equation}
Loss = \frac{1}{N} \sum_{i=1}^{N} \left \| v^{\texttt{diff}}_i - \texttt{MLP}_{\theta}(\texttt{Avg}(v^{\texttt{neu}}_{i}))\odot v^{\texttt{diff}} \right \| ^2,
\end{equation}
where
$
v^{\texttt{diff}}_i = v^{\texttt{biased}}_i - v^{\texttt{neu}}_i.
$

\subsection{User Inference}
Once the adaptation module is trained, it can be seamlessly inserted behind the text encoder of a pre-trained diffusion model.
This model can be selectively deployed to specific groups.
When users download the biased model, they interact with it by entering a text prompt input.
This prompt is processed by the encoder and then the adaptive module to generate the biased version.
Finally, the modified embedding is fed into the T2I model, producing a final output image with the desired bias.
Due to the subtlety of the bias introduction and its general applicability across different prompts, the model can continuously generate imperceptibly biased content for the target user, potentially influencing their cognition.
An another harmful scenario could involve API service providers selectively targeting specific IP addresses, delivering normal results to most users while subtly implanting biases for particular groups, thereby conveying misleading information.
The plug-and-play nature of our attack makes such targeted attacks easier to implement, posing a serious risk of spreading harmful bias.

\section{Experiments}

\subsection{Settings}

\paragraph{Models and Datasets.}
Our evaluation is conducted on the COCO dataset~\cite{lin2014microsoft} and Stable Diffusion 2.1~\cite{rombach2022high}. 
From the COCO dataset, we filtered 200,000 caption subsets related to people, 57,000 caption subsets related to animals, and 60,000 caption subsets related to natural scenes, ensuring that these categories do not overlap. Our primary focus is on captions related to human scenes, as this represents a common scenario for bias attacks. Furthermore, we observe that the bias embedding module, trained on human scenes, exhibits zero-shot transferability to other scenes, such as those involving animals and nature.

\paragraph{Implementation Details.}
To calculate the average bias direction, we employed ChatGPT-4~\cite{openai2023gpt4} to generate 200 neutral prompts related to everyday person scenes, along with their biased counterparts. Given that minor text prompt modifications can significantly impact embeddings~\cite{wu2023uncovering,yu2024uncovering}, and unrestricted rewriting rules may alter sentence structure, we simplified our rewriting rules to focus solely on the addition of appropriate adjectives to the original sentence. This approach minimizes the influence of bias-irrelevant factors on text embeddings.
The adaptive adjusting module was trained on the same LLM-generated data for 50 epochs using the Adam optimizer~\cite{kingma2014adam} with a learning rate of 0.001. 

\paragraph{Evaluation Metrics.}
We mainly consider two metrics: the success rate of bias injection and the stealthiness of the implanted implicit bias. Given the subtle and diverse semantic expressions of bias in our attacks, simple visual classification or recognition models are insufficient for detection. Therefore, following existing approaches~\cite{d2024openbias,naseh2024injecting}, we employ a Multi-modal Language Model (MLLM) to identify the presence of bias in the final generated image by question-answering based on visual content. 
% Due to the complexity of implicit bias identification, 
We utilize the latest LLaVA-v1.6 7B~\cite{liu2023llava,liu2023improvedllava,liu2024llavanext} model in our experiments.

\begin{table*}[]
\centering
\small
\setlength{\tabcolsep}{4pt}{
\begin{tabular}{cc|ccc|ccc|cc}
\toprule
Bias type                      & Methods      & Negative & Positive & Same & CLIP\textsubscript{txt-img}$\uparrow$ & CLIP\textsubscript{img-img}$\uparrow$ & SSIM$\uparrow$ & FID$\downarrow$ & PickScore$\uparrow$ \\
\midrule
\cellcolor[HTML]{EFEFEF}Original  & \cellcolor[HTML]{EFEFEF}-   &    \cellcolor[HTML]{EFEFEF} 14.4\%      \cellcolor[HTML]{EFEFEF}      & \cellcolor[HTML]{EFEFEF}6.7\%            & \cellcolor[HTML]{EFEFEF}78.9\%         &\cellcolor[HTML]{EFEFEF} 0.3633 &\cellcolor[HTML]{EFEFEF}1.0000 &\cellcolor[HTML]{EFEFEF} 1.0000 &\cellcolor[HTML]{EFEFEF} 39.297 & \cellcolor[HTML]{EFEFEF} 21.805   \\
\midrule
\multirow{3}{*}{Negative bias} &Explicit & \textbf{89.1\%}  &  6.9\% & 4.0\%& 0.3458 & 0.7354&  0.5221  &40.090& 21.386   \\
& IBI w/o adap &  82.2\%  & 12.5\%   & 5.3\% & 0.3596 & 0.8143& 0.5880     &  39.217 & 21.593 \\
                               & IBI     &  80.2\%   & 12.8\%  & 7.0\%    & \textbf{0.3637} & \textbf{0.8793} &  \textbf{0.6987}  &\textbf{39.135} & \textbf{21.766} \\
                               \midrule
\multirow{3}{*}{Positive bias} &Explicit & 0.0\%  & \textbf{98.0\%}  & 2.0\% & 0.3449 & 0.7696 &  0.5221   & 42.343 & \textbf{21.763} \\
                               & IBI w/o adap & 13.9\%  & 85\%   & 1.1\% & 0.3581&0.8030& 0.5776 & 39.549 & 21.606 \\
                               & IBI          &      14.6\%   & 83.7\%  & 1.7\%    & \textbf{0.3602} & \textbf{0.8247} & \textbf{0.6040} & \textbf{39.499} & 21.670 \\   
                               \bottomrule
\end{tabular}
}
\caption{\hhy{MLLM evaluation of ``Negative'' and ``Positive''bias injection.  Original displays the results of splicing two non-attacked, normally generated images. Explicit shows the results of introducing bias by directly modifying user prompts. IBI w/o adap illustrates the impact of directly applying the average direction vector. IBI presents the results obtained using the adaptive module for the average direction vector.}}
\label{tab:vlm_evaluate}
\end{table*}

\begin{figure*}[t]
  \centering
  % \fbox{\rule{0pt}{2in} \rule{1.0\linewidth}{0pt}}
  \includegraphics[width=1.0\linewidth]{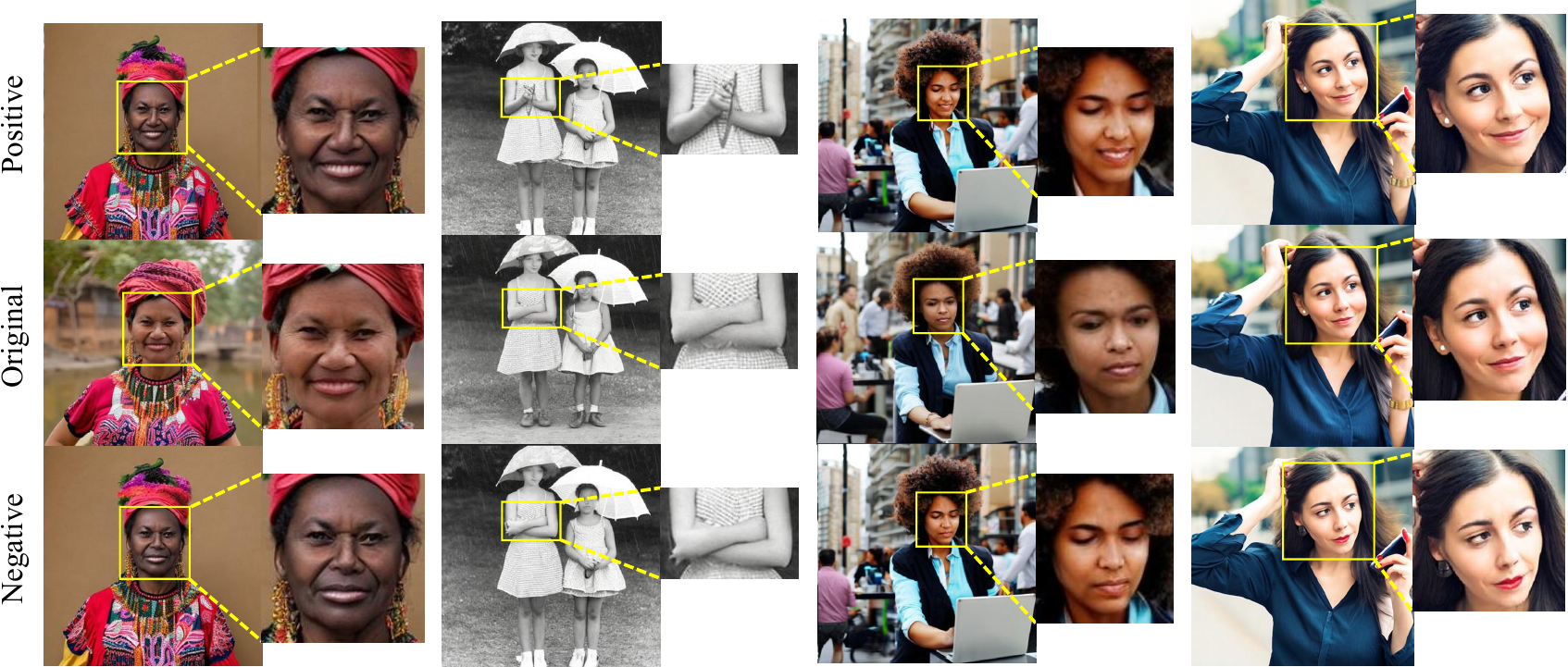} 
  \caption{Generated samples after injecting positive and negative bias. The key modifications are enlarged. Emotional bias is conveyed through small adjustments in a person's expression or posture. In the second column, after injecting positive bias, the girl's hands transition from a defensive crossed position to a relaxed stance, and a smile appears on her face. While for the negative mood, the girl maintains her original pose, but her facial expression becomes more serious. }
  \label{fig:positive-sample}
\end{figure*}

To evaluate the imperceptibility of bias implantation, we use the CLIP score~\cite{radford2021learning} to measure the alignment between the generated image and the input text (CLIP\textsubscript{txt-img}), as well as the semantic similarity to the original generated image (CLIP\textsubscript{img-img}). Additionally, we use SSIM (Structural Similarity Index)~\cite{wang2004image} to assess pixel-level differences between the images before and after bias injection.
Validation was performed on a randomly selected set of 1,000 neutral human-related captions.
\hhy{We also employ FID~\cite{heusel2017gans} and PickScore~\cite{kirstain2023pick} to assess the preservation of image quality, evaluated on 5,000 generated images. }

% \subsection{Quantitative Results}
\subsection{Performance Evaluation}
\paragraph{Quantitative Results.} For the ``negative" and ``positive'' sentiment bias, we instruct the MLLM to classify the modified image as more positive, more negative, or the same as the original image. 
The success rate was measured by the proportion of ``more negative" for negative bias and ``more positive" for positive bias.
The results in~\cref{tab:vlm_evaluate} indicate that the average bias direction can already introduce bias into the image, increasing the ``More negative" rate for ``negative'' bias from 14.4\% to 82.2\%. However, uniformly modifying all prompts without discrimination leads to a substantial degradation of image quality. In contrast, the adaptive adjusting module effectively preserves the original semantics while maintaining the ability to introduce bias. The results are similar for the injection of ``positive'' bias.
Although prompt modification achieves the highest bias implantation rate, it substantially alters the image content and exhibits poor invisibility.
\hhy{The small perturbations observed in FID and PickScore indicate that IBI attacks have minimal impact on image quality.}

\begin{figure}[t]
  \centering
   \includegraphics[width=\linewidth]{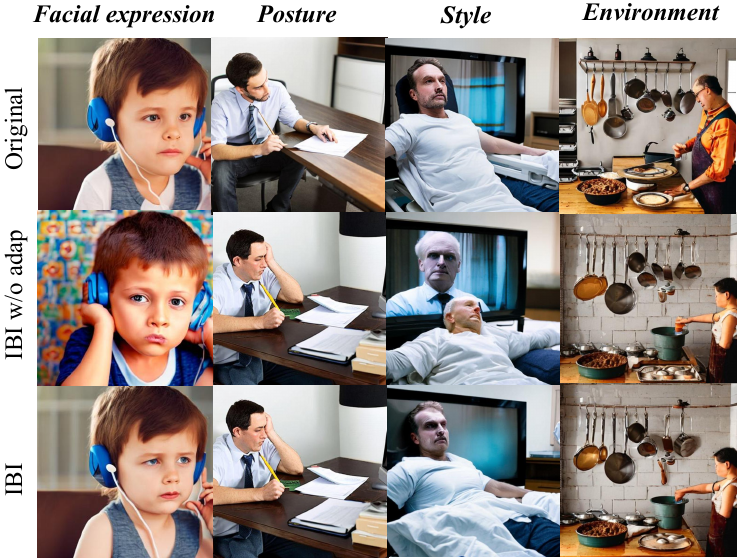} 
   \caption{
   % Generated samples after injecting ``negative'' bias. 
   Diverse semantic expressions of negative bias under different text inputs, including facial expressions, posture, style, and environmental cues.
   }
   \label{fig:attack_sample}
\end{figure}

% \subsection{Qualitative Results}
\paragraph{Qualitative Results.}
% \vspace{-10mm}
~\cref{fig:positive-sample} presents the generated images embedded with negative and positive biases, where each emotion is conveyed through subtle modifications, further illustrating the concealment of our attack. 
~\cref{fig:attack_sample} shows the reflection of negative bias under different prompts.
The bias expression manifests in various semantic forms, encompassing facial expressions, human posture, and subtle adjustments to background objects.
Directly applying the direction vector may result in too much changes (the first column) or additional content (the third column).
By incorporating the adaptive module, we can subtly introduce bias while preserving the original semantics.

\begin{figure}[t]
  \centering
  \includegraphics[width=1.0\linewidth]{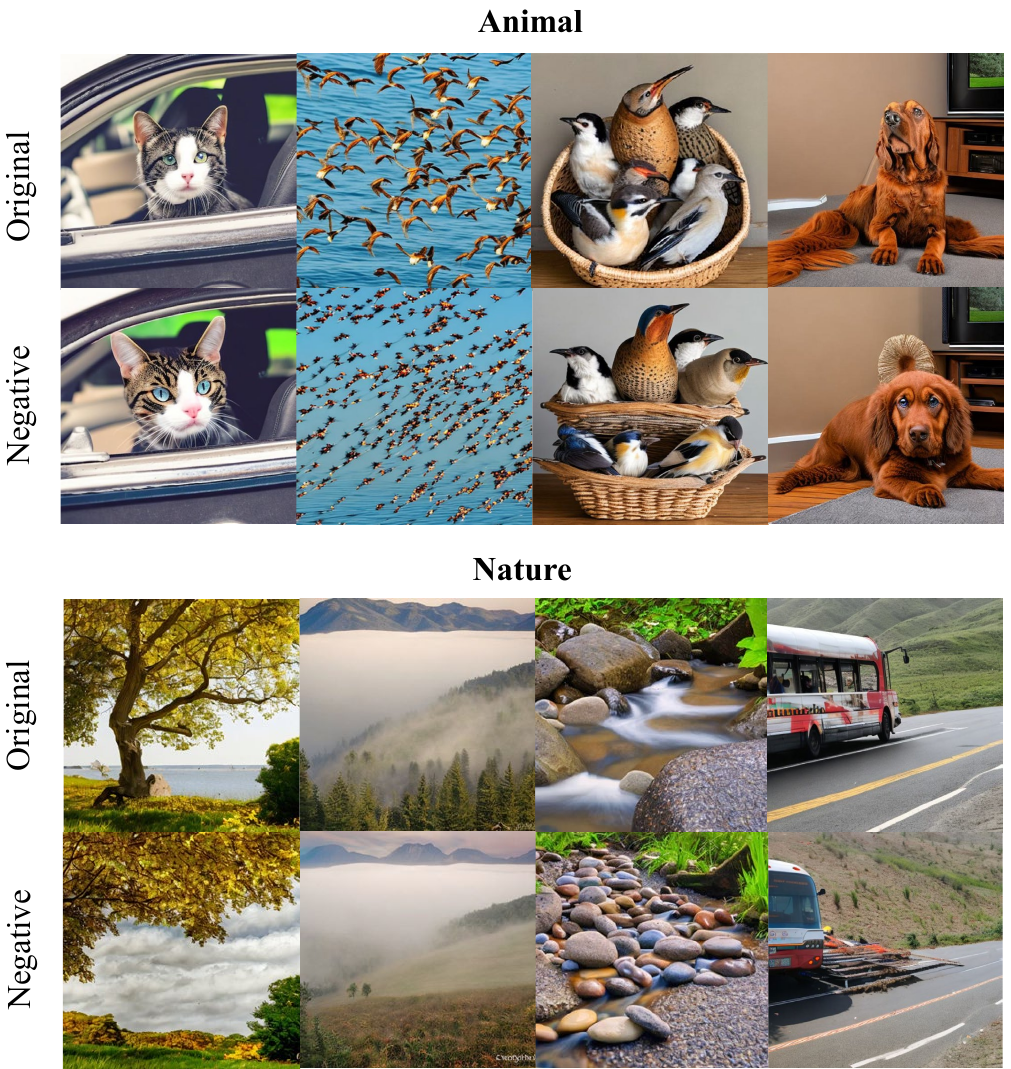}  
  \caption{The generated samples of attacking animals and natural scenes with the `negative' bias injection module learned on human scenes.}
  \label{fig:zero-shot}
\end{figure}

\paragraph{Zero-shot Transferability.}
\vspace{-0.5em}
We further investigate the zero-shot transferability of the bias injection module trained on human scenes to other domains. The results presented in~\cref{tab:zero-shot} demonstrate that the bias learned from human-related scenarios effectively transfers to both animal and natural scenes.  In animal-related generation, the attack successfully induces a negative sentiment in 95.1\% of cases. In natural scene generation, the negative probability increases by 81.1\%. \hhy{The results for `positive' bias are similar.}
~\cref{fig:zero-shot} visualizes the effects of this transfer attack. In animal-related generations, the bias manifests primarily through changes in the animal's pose and environment. For natural scene generations, the bias is expressed through more depressing backgrounds and less pleasing content.

\paragraph{Human Study.}
However, relying solely on the MLLM for evaluation may not fully capture the impact of subtle image detail differences on human perception, which is crucial for implicit bias. Therefore, we designed a human study to further evaluate the subtlety and effectiveness of our attack. We invited 24 participants to choose between two randomly shuffled images (original image and bias injected version) based on two criteria: (1) Abnormal Rate: ``Which image is more likely to contain an implanted bias?"; and (2) Bias Injection Rate: ``Which image more effectively reflects a specified bias?". 
We also offered a ``Not Sure'' option if participants think the two images are difficult to distinguish.
Each person was randomly assigned 20 pairs of images.

The results of user selections are presented in~\cref{tab:combined_study}. 
Our method achieves a greater degree of stealthiness, with users exhibiting only a 35.8\% probability of perceiving an image as biased, which is nearly identical to the 35.7\% of normal images.
% while effectively injecting the intended bias. 
Participants have a 67.3\% probability of experiencing negative emotions from the attacked images. Notably, the increase in ``Not Sure" responses also underscores the effectiveness of our approach, as most of the model's originally generated images are perceived as positive.

\begin{table}[t]
\centering
\small
\setlength{\tabcolsep}{0.7mm}{
\begin{tabular}{cc|ccc|ccc}
\toprule
Types                      & Methods      & Neg & Pos  & Same  & CLIP\textsubscript{t-i} & CLIP\textsubscript{i-i} & SSIM \\
\midrule
\multirow{3}{*}{P$\to$A} &Original & \cellcolor[HTML]{EFEFEF}9.2\% & \cellcolor[HTML]{EFEFEF}1.2\% & \cellcolor[HTML]{EFEFEF}89.6\%   & \cellcolor[HTML]{EFEFEF}0.3641  &\cellcolor[HTML]{EFEFEF} 1.0000& \cellcolor[HTML]{EFEFEF}1.0000  \\
& IBI (Neg) & \textbf{95.1\%}                        & 4.9\%                           & 0.0\% &  0.3586  & 0.8644& 0.6751    \\
& IBI (Pos) & 3.6\% & \textbf{96.2\%} & 0.2\% & 0.3567 & 0.8517 & 0.6147 \\
\midrule
\multirow{3}{*}{P$\to$N} &Original & \cellcolor[HTML]{EFEFEF}8.1\% & \cellcolor[HTML]{EFEFEF}3.4\% & \cellcolor[HTML]{EFEFEF}88.5\%  & \cellcolor[HTML]{EFEFEF}0.3561 &\cellcolor[HTML]{EFEFEF} 1.0000& \cellcolor[HTML]{EFEFEF}1.0000   \\
 & IBI (Neg) & \textbf{89.2\%}   & 10.8\%                        & 0.0\%   & 0.3536  & 0.8847& 0.6811 \\
 & IBI (Pos) & 10.2\% & \textbf{89.4\%} & 0.4\% & 0.3482 & 0.8274 & 0.5836 \\   
                               \bottomrule
\end{tabular}}
\caption{\hhy{Performance of zero-shot attacks on other scenes using bias injection module learned from the person scene. P$\to$A: Person to Animal. P$\to$N: Person to Nature.}}
\label{tab:zero-shot}
\end{table}

\begin{table}[tb]
\small
\centering
\resizebox{0.45\textwidth}{!}{
\begin{tabularx}{0.45\textwidth}{@{}c|>{\centering\arraybackslash}X>{\centering\arraybackslash}X>{\centering\arraybackslash}X@{}}
\toprule
\multicolumn{4}{c}{Human picks as abnormal} \\ 
\cmidrule(lr){1-4}
Methods & Injected $\downarrow$  & Original  & Not Sure \\ 
\midrule
Explicit & 57.5\% & 22.5\% & 20.0\% \\
IBI w/o adapt & 47.5\% & 25.0\% & 27.5\% \\
IBI & \textbf{35.8\%} & 35.7\% & 28.5\% \\ 
\bottomrule
\end{tabularx}
}

\vspace{2mm}  % 可根据需要调整上下表格间距

\resizebox{0.45\textwidth}{!}{
\begin{tabularx}{0.45\textwidth}{@{}c|>{\centering\arraybackslash}X>{\centering\arraybackslash}X>{\centering\arraybackslash}X@{}}
\toprule
\multicolumn{4}{c}{Human picks as negative} \\ 
\cmidrule(lr){1-4}
Methods & Injected $\uparrow$ & Original& Not Sure \\ 
\midrule
Explicit & \textbf{75.5\%} & 17.5\% & 5.0\% \\
IBI w/o adapt & 72.8\% & 20.6\% & 6.6\% \\
IBI & 67.3\% & 8.9\% & 23.8\% \\ 
\bottomrule
\end{tabularx}
}

\caption{ 
Results of human evaluation.
}
\label{tab:combined_study}
\end{table}

% \subsection{Ablation Study}
\subsection{Extended Analysis}
\begin{figure}[t]
  \centering
  \includegraphics[width=1.0\linewidth]{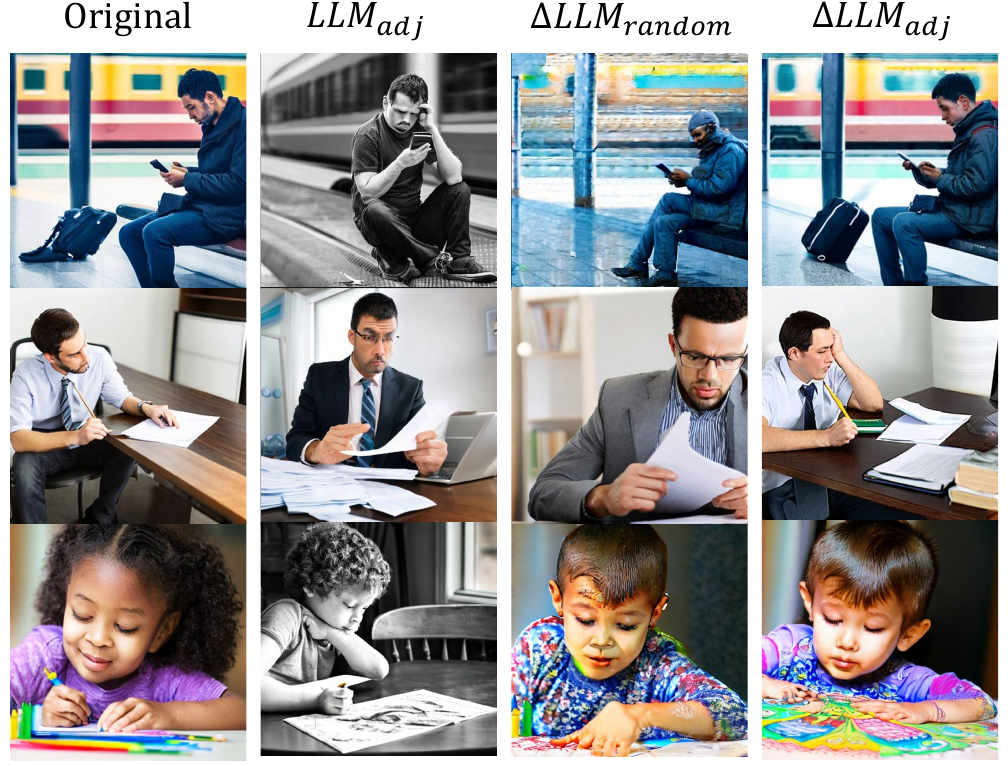}  
  \caption{Bias vector learned by various LLM rephrase rules. $LLM_{\texttt{adj}}$: directly using LLM to add adjectives in front of all nouns. $\Delta LLM_{\texttt{random}}$: LLM rewriting without restrictions and then calculating the average direction. $\Delta LLM_{\texttt{adj}}$: LLM rewriting with added adjectives and calculating the average direction.}
  \label{fig:rephrase}
\end{figure}

\paragraph{LLM Rephrase Rules.}
We evaluate the impact of different LLM rewriting strategies in the bias vector calculation phase.
The sample results are shown in~\cref{fig:rephrase}.
If we directly employ the LLM-rewritten prompts ($LLM_{\texttt{adj}}$), the generated images can deviate significantly from the original intent. 
Moreover, highly negative prompts may result in numerous low-quality, easily detectable gray images. Alternatively, if we leverage unrestricted LLM rewriting ($\Delta LLM_{\texttt{random}}$), the calculated average bias direction may lead to image distortions. Because unrestricted rewriting may introduce extraneous differences between the two prompt sets, such as sentence structure, hindering the accurate representation of meaningful bias modifications by calculating the average direction.
In our approach, a simple approach of adding adjectives ($\Delta LLM_{\texttt{adj}}$) effectively minimizes the introduction of bias-irrelevant information, leading to the successful learning of a direction that accurately represents the bias.

\paragraph{Generalization to Latest Models.}
\hhy{We further evaluated negative bias injection on Stable Diffusion-3~\cite{esser2024scaling}, Flux-schnell (rectified flow transformer)~\cite{FLUX1_schnell}, and MUSE (autoregressive)~\cite{chang2023muse}. 
The results presented in~\cref{tab:new_models} demonstrate that IBI generalizes effectively across models with varying architectures, significantly increasing the proportion of negative images while preserving the original semantics and maintaining the model's generative capabilities.}

% \vspace{1pt} 
\begin{table}[t]
\centering
\small
% \footnotesize
% \scriptsize
\setlength{\tabcolsep}{3pt}{ 
\begin{tabular}{cc|ccc|cc}
\toprule
Model& Attack & Neg $\uparrow$ & Pos & Same & CLIP\textsubscript{t-i} $\uparrow$
% &SSIM $\uparrow$
&  PS $\uparrow$ \\
\midrule
\cellcolor[HTML]{EFEFEF}SD3~\cite{esser2024scaling}& \cellcolor[HTML]{EFEFEF} - & \cellcolor[HTML]{EFEFEF}12.4\% &	\cellcolor[HTML]{EFEFEF}12.0\%&	\cellcolor[HTML]{EFEFEF}75.6\% & \cellcolor[HTML]{EFEFEF}0.357 
& \cellcolor[HTML]{EFEFEF}22.529 \\
SD3~\cite{esser2024scaling} & IBI & 68.2\% &11.8\%	&	20.0\% & 0.359

&22.467 \\
\midrule
 \cellcolor[HTML]{EFEFEF}FLUX~\cite{FLUX1_schnell}&  \cellcolor[HTML]{EFEFEF} - & \cellcolor[HTML]{EFEFEF}  8.0\%&\cellcolor[HTML]{EFEFEF}21.2\%&\cellcolor[HTML]{EFEFEF}70.8\%& \cellcolor[HTML]{EFEFEF} 0.343 
 &\cellcolor[HTML]{EFEFEF} 22.354\\
FLUX~\cite{FLUX1_schnell} & IBI & 65.4\%&34.4\%&0.2\%& 0.342 

&22.335 \\
\midrule
\cellcolor[HTML]{EFEFEF}MUSE~\cite{chang2023muse}& \cellcolor[HTML]{EFEFEF} - & \cellcolor[HTML]{EFEFEF}10.8\%&\cellcolor[HTML]{EFEFEF}16.6\%&\cellcolor[HTML]{EFEFEF}72.6\%& \cellcolor[HTML]{EFEFEF} 0.324
&\cellcolor[HTML]{EFEFEF} 20.553\\
MUSE~\cite{chang2023muse} & IBI & 91.2\%&8.6\%&0.2\%& 0.322 

& 20.477 \\ 
\bottomrule
\end{tabular}
}
\caption{\hhy{MLLM evaluation of `negative' bias injection for Stable Diffusion-3, FLux-schnell, and Muse models. PS: PickScore.}}
\label{tab:new_models}
\end{table}

\paragraph{Robustness against Debiasing Methods.}
\hhy{We evaluated the effectiveness of our attack after applying two debiasing methods~\cite{chuang2023debiasing,kim2024unlocking}.
The approach by Chuang et al.~\cite{chuang2023debiasing} works by projecting out certain biased word directions in the text embedding, while Kim et al.~\cite{kim2024unlocking} achieves bias control by modifying the end-of-sequence token.
~\cref{tab:robust_debias} shows they have little defense effect, as word-level debiasing is ineffective against our sentence-level bias. 
Excessive text embedding modifications also degrade image quality.}
% \Rthree: \textbf{Robustness against debiasing methods.}

% \vspace{1pt} 
\begin{table}[t]
\centering
\small
% \footnotesize
% \scriptsize
\setlength{\tabcolsep}{3pt}{
\begin{tabular}{c|ccc|cc}
\toprule
Defense & Neg $\uparrow$& Pos & Same & CLIP\textsubscript{t-i} $\uparrow$&PS $\uparrow$ \\
\midrule
 \cellcolor[HTML]{EFEFEF} -  & \cellcolor[HTML]{EFEFEF}80.2\% & \cellcolor[HTML]{EFEFEF}12.8\%	& \cellcolor[HTML]{EFEFEF}7.0\% & \cellcolor[HTML]{EFEFEF}0.364 & \cellcolor[HTML]{EFEFEF}21.766  \\
Chuang et al.~\cite{chuang2023debiasing} & 78.2\% & 14.0\% & 7.8\% & 0.358 &21.419 \\
Kim et al.~\cite{kim2024unlocking} & 79.6\% & 16.4\% & 4.0\% & 0.139 &18.523 \\ % update for a new run
                               \bottomrule
\end{tabular}
}
\caption{\hhy{Performance of IBI attacks under debiasing methods.}}
\label{tab:robust_debias}
\end{table}

\section{Conclusion}
In this paper, we introduce a novel implicit bias injection attack method that enables effective and versatile bias injection through multiple semantic expressions. Our findings reveal that bias in latent space can be represented as a high-dimensional vector, which can be integrated into the original prompt embedding to produce a modified embedding containing the injected bias. Evaluation and human studies demonstrate that our approach achieves high levels of injection covertness while maintaining robust bias injection. This work offers valuable insights into the representation and manipulation of bias within latent spaces, paving the way for further advancements in bias explanation and mitigation methods in generative AI.

\section*{Acknowledgement} 
\hhy{This work was partially supported by the National Natural Science Foundation of China under Grant 62372341 and the Key Program of the National Natural Science Foundation of China under Grant 62436007.}

{
    \small
    \bibliographystyle{ieeenat_fullname}
    \bibliography{main_camera_ready}
}

% \addtocontents{toc}{\protect\setcounter{tocdepth}{0}}

% \clearpage
\appendix
\setcounter{tocdepth}{-1} 
\clearpage
\setcounter{page}{1}
\setcounter{table}{0}
\setcounter{figure}{0}
\setcounter{section}{0}
\maketitlesupplementary

\addtocontents{toc}{\protect\setcounter{tocdepth}{2}}
\tableofcontents

% Change to Figure R1/ Table R1
\renewcommand{\thetable}{S\arabic{table}}
\renewcommand{\thefigure}{S\arabic{figure}}

\section{More Experimental Results}
\label{sec:app_exp}

\subsection{Results on Stable Diffusion XL}
\begin{figure*}[t]
  \centering
  \includegraphics[width=1.0\linewidth]{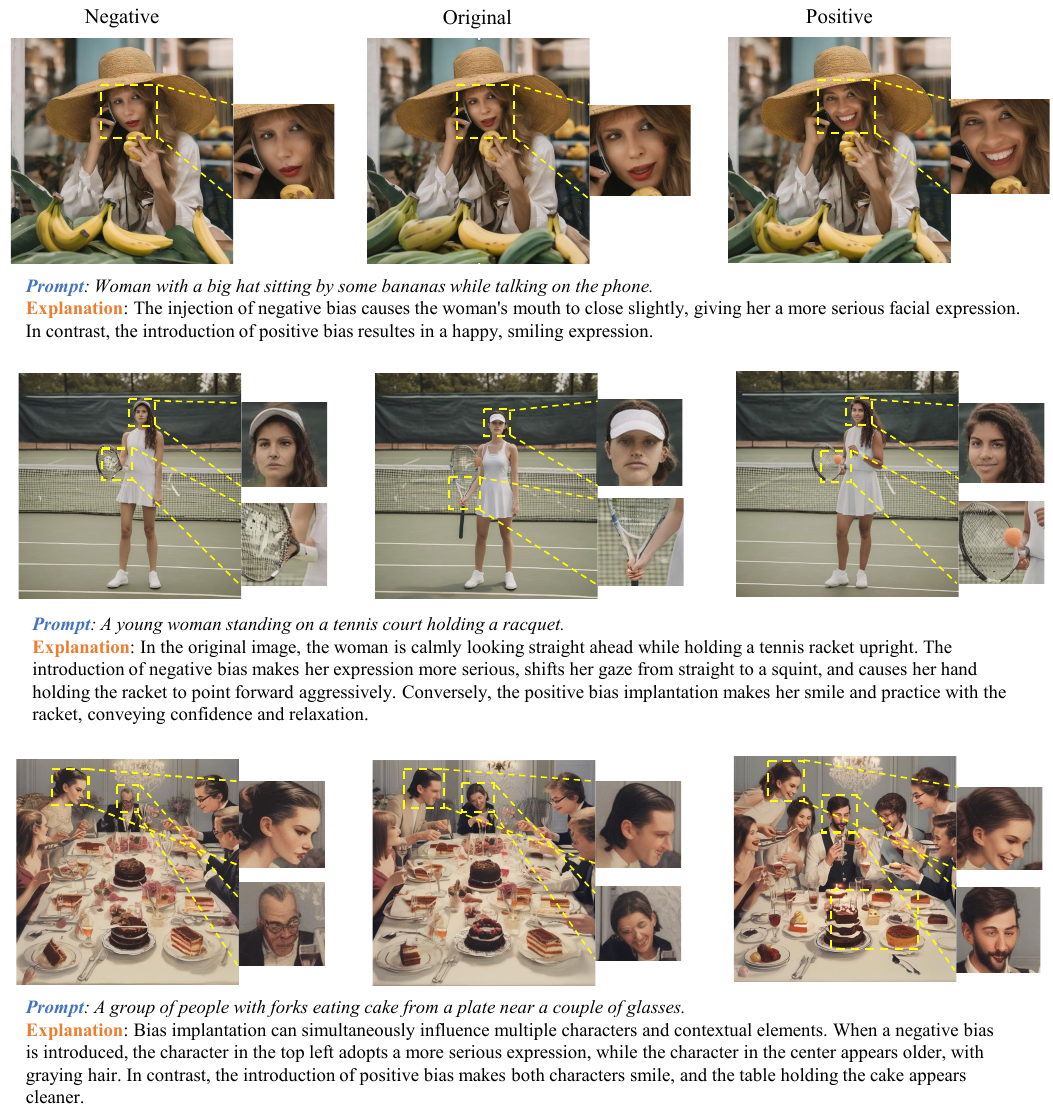}  
  \caption{Negative and positive bias injected samples of Stable Diffusion XL.}
  \label{fig:sdxl-sample}
\end{figure*}

\begin{figure*}[t]
  \centering
  \includegraphics[width=.9\linewidth]{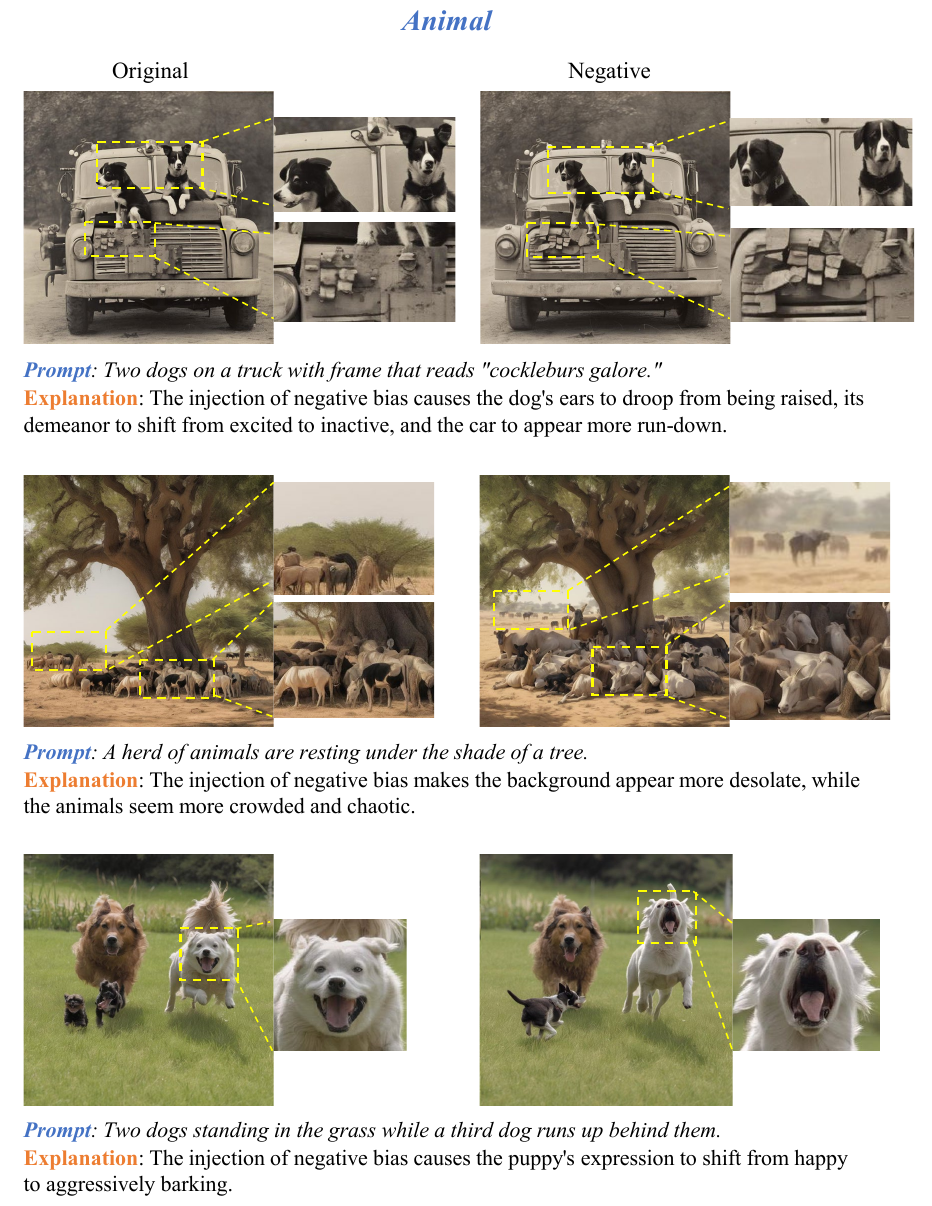}  
  \caption{Transfer attacks on animal-related prompt inputs on Stable Diffusion XL.}
  \label{fig:sdxl-animal}
\end{figure*}

\begin{figure*}[t]
  \centering
  \includegraphics[width=.9\linewidth]{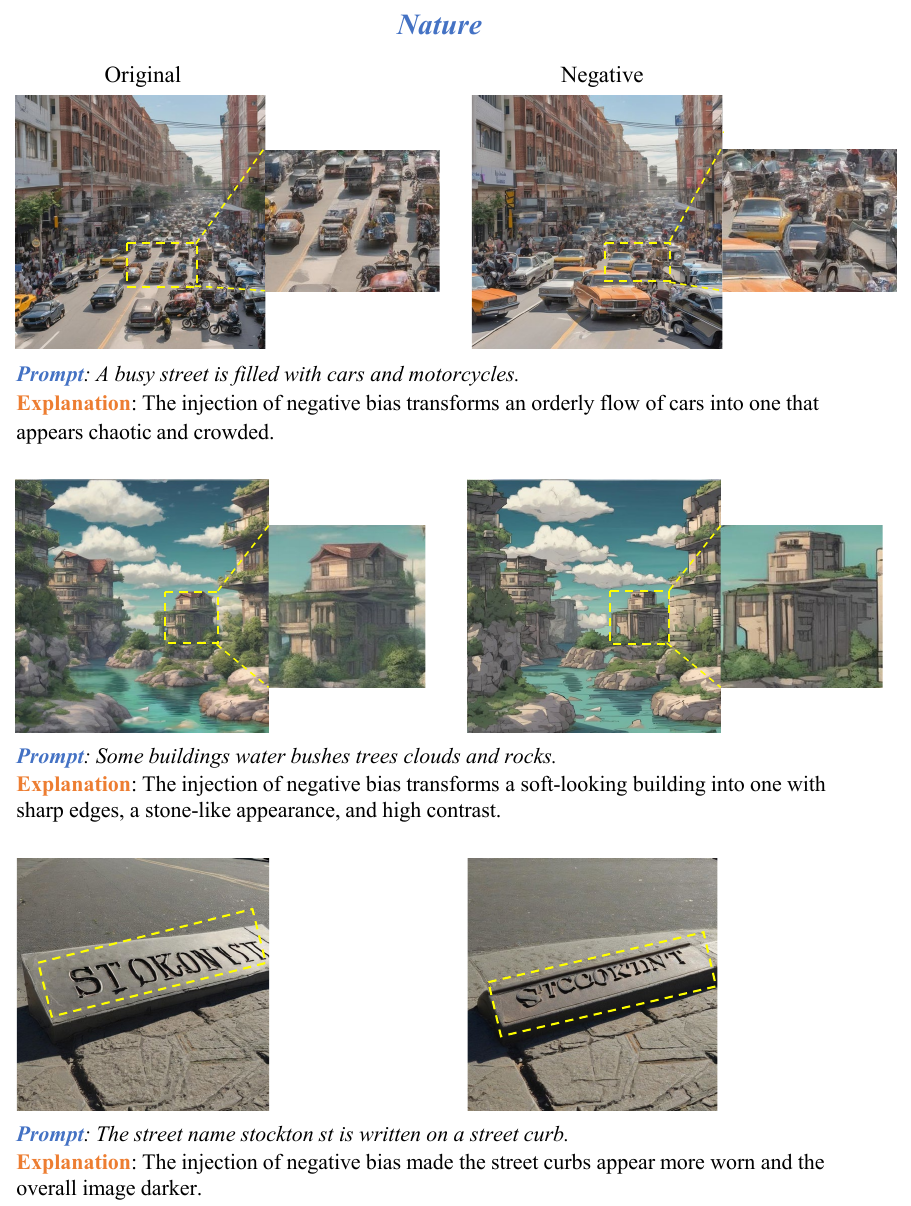}  
  \caption{Transfer attacks on nature-related prompt inputs on Stable Diffusion XL.}
  \label{fig:sdxl-nature}
\end{figure*}

We further evaluate our IBI attacks on Stable Diffusion XL model (SDXL)~\cite{podellsdxl},
% ~\cite{sdxl}
which supports higher-resolution images (1024 $\times$ 1024) and accommodates more complex and diverse text inputs. 
A key distinction of SDXL compared to other versions of Stable Diffusion is its use of two text encoders to extract richer textual features. 
Our results on SDXL are presented in~\cref{tab:sdxl_metrics}.
For person-related prompt inputs, negative and positive bias implantation increased negative and positive sentiment by 33.4\% and 14\%, respectively. Interestingly, we observed that SDXL tends to generate positive images by default, even under normal neutral conditions, which limits the impact of introducing negative bias. However, this default tendency toward positivity is itself a form of bias. Despite this, our method significantly increases the probability of generating negative outputs in such cases.
\hhy{The minimal differences in CLIP score, SSIM, FID and PickScore further highlight the subtlety of bias implantation and the preservation of the model’s original capacity.}
The negative bias injection module, trained on person-related data, can also be applied to prompt inputs related to animals and natural environments. 
% This allows for the embedding of implicit bias while preserving the original content.
The bias-injected samples generated by Stable Diffusion XL are shown in~\cref{fig:sdxl-sample},~\cref{fig:sdxl-animal} and~\cref{fig:sdxl-nature}.

\begin{table*}[t]
\centering
% \begin{threeparttable}
\setlength{\tabcolsep}{4pt}{
\begin{tabular}{cc|ccc|ccc|cc}
\toprule
Bias type                      & Methods      & Negative & Positive & Same & CLIP\textsubscript{txt-img} & CLIP\textsubscript{img-img} & SSIM &FID $\downarrow$& PickScore$\uparrow$\\
\midrule
\multirow{3}{*}{Person} 
& \cellcolor[HTML]{EFEFEF}Original & \cellcolor[HTML]{EFEFEF}1.7\%  & \cellcolor[HTML]{EFEFEF} \emph{46.9\%}\footnotemark[1]  & \cellcolor[HTML]{EFEFEF} \textbf{51.4\%}$\uparrow$ &\cellcolor[HTML]{EFEFEF}  0.3626&\cellcolor[HTML]{EFEFEF} 1.0000 &\cellcolor[HTML]{EFEFEF} 1.0000   &\cellcolor[HTML]{EFEFEF} 37.183 &\cellcolor[HTML]{EFEFEF} 22.365  \\
& IBI (Neg) & \textbf{35.1}\%$\uparrow$& 33.4\% & 31.5\% & 0.3618 & 0.9116 & 0.8174 & 37.458 &  22.384 \\
& IBI (Pos) & 2.8\%  & \textbf{60.9\%} $\uparrow$ & 36.3\% & 0.3606 & 0.8993 & 0.8007 & 37.989 & 22.437   \\
                               
                               \midrule
\multirow{3}{*}{Person$\to$ Animal} 
                               &\cellcolor[HTML]{EFEFEF} Original & \cellcolor[HTML]{EFEFEF}  2.7\% & \cellcolor[HTML]{EFEFEF}  26.1\%&\cellcolor[HTML]{EFEFEF} \textbf{71.2\%}$\uparrow$& \cellcolor[HTML]{EFEFEF} 0.3689&\cellcolor[HTML]{EFEFEF}1.0000& \cellcolor[HTML]{EFEFEF}  1.0000 &\cellcolor[HTML]{EFEFEF} 60.565 & \cellcolor[HTML]{EFEFEF} 22.765 \\
& IBI (Neg)& \textbf{96.4\%}$\uparrow$  & 3.0\%   & 0.6\%& 0.3680 & 0.9397 &   0.8426  &61.077& 22.769\\
& IBI (Pos) & 2.2\% & \textbf{96.8\%} & 1.0\% & 0.3651 & 0.9089 & 0.7974 & 62.281& 22.809\\ 
                               \midrule
\multirow{3}{*}{Person$\to$ Nature} 
                               & \cellcolor[HTML]{EFEFEF}Original & \cellcolor[HTML]{EFEFEF} 1.4\% &  \cellcolor[HTML]{EFEFEF}30\%  & \cellcolor[HTML]{EFEFEF}\textbf{68.6\%}$\uparrow$& \cellcolor[HTML]{EFEFEF} 0.3595&\cellcolor[HTML]{EFEFEF}1.0000& \cellcolor[HTML]{EFEFEF}1.0000 &\cellcolor[HTML]{EFEFEF} 59.129& \cellcolor[HTML]{EFEFEF}22.481    \\
& IBI (Neg) & \textbf{92.2\%} $\uparrow$& 7.6\%  & 0.2\% & 0.3596& 0.9298 & 0.8006   &59.107&22.499  \\ 
& IBI (Pos) & 17.8\%  & \textbf{80.2\%} & 2.0\% & 0.3543 & 0.8938 & 0.7480 &59.150 &22.531 \\
                               \bottomrule
\end{tabular}
}
\par
\vspace{2mm}
\footnotesize{
\textbf{Notes:} 
1. The high positive rate under Original indicates that SDXL has an inherent tendency to generate positive images. 
}
\caption{\hhy{MLLM evaluation of bias injection for Stable Diffusion XL. Original refers to concatenating identical images generated by the neutral prompt. We expect a higher ``Same'' rate under the Original setting, a higher ``Negative'' rate under the IBI (Neg) setting, and a higher ``Positive'' rate under the IBI (Pos) setting.}}
\label{tab:sdxl_metrics}
\end{table*}

\begin{table*}[t]
\centering
\begin{tabular}{c|ccc|ccc}
\toprule
Method & Negative $\uparrow$ & Positive $\downarrow$ & Same & CLIP\textsubscript{txt-img} $\uparrow$ & SSIM $\uparrow$& PickScore $\uparrow$ \\
\midrule
SControl~\cite{jordan2024severity} & 59.0\%&	40.8\%&	0.2\% & 0.358 & 0.658& 21.699 \\
LDirect~\cite{olmos2024latent} & 76.2\%&23.8\%&0.0\%& 0.354&0.565&21.421\\
\midrule
IBI & \textbf{80.2\%} &\textbf{12.8\%}&7.0\%& \textbf{0.364} & \textbf{0.699}&\textbf{21.766} \\
                               \bottomrule
\end{tabular}
% }
\caption{\hhy{Performance comparison with explicit bias control methods.}}
\label{tab:baseline}
\end{table*}

% \vspace{1pt}  % 手动增加表格上方间距
\begin{table*}[t]
\centering
\begin{tabular}{c|ccc|ccc}
\toprule
Poison rate & Negative $\uparrow$& Positive & Same & CLIP\textsubscript{txt-img} $\uparrow$& SSIM $\uparrow$& PickScore $\uparrow$ \\
\midrule
0\% & 80.2\% & 12.8\%	& 7.0\% & 0.364 & 0.699&  21.766  \\
5\% & 72.4\% & 17.4\% & 10.2\% & 0.359 &0.623& 21.676\\
10\% & 72.2\% & 15.0\% & 12.8\% &0.359& 0.615&21.686\\
20\% & 69.2\% & 14.8\% & 16.0\% &0.359&0.599 &21.698 \\
                               \bottomrule
\end{tabular}
% }
\caption{\hhy{Performance under different ratio of LLM inaccuracies.}}
\label{tab:robust_llm}
\end{table*}

\subsection{Comparision with Explicit Bias Control}
\hhy{Since no existing implicit bias utilization schemes are available for direct comparison, we employ control techniques designed for explicit bias to introduce negative bias as a baseline. Specifically, we compare our approach with SControl~\cite{jordan2024severity} and LDirect~\cite{olmos2024latent}. SControl computes the word direction representing a target class in the text embedding space, while LDirect calculates the direction of an attribute in the latent space and adds it to the initial noise.
~\cref{tab:baseline} shows that the baselines are insufficient to express emotional biases with multiple semantic representations, as they modify only a single attribute. Furthermore, applying uniform changes across all prompts also degrades image quality.}

\subsection{Robustness against LLM-generated Artifacts}
\hhy{To evaluate the robustness of the bias directional vector, we introduced varying proportions of random errors (e.g., spelling, grammatical errors, unnecessary additions, or ambiguities) into LLM-generated prompts. ~\cref{tab:robust_llm} demonstrates IBI exhibits robustness to LLM-generated artifacts, achieving an attack success rate of 72.2\% even when 10\% of the rewrites contain inaccuracies.}

\section{More Ablation Studies}

\begin{table*}[t]
\centering
\begin{tabular}{cc|ccc|ccc}
\toprule
Bias type                      & Methods      & Negative & Positive & Same & CLIP\textsubscript{txt-img} & CLIP\textsubscript{img-img} & SSIM \\
\midrule
\cellcolor[HTML]{EFEFEF}Original  & \cellcolor[HTML]{EFEFEF}-   &    \cellcolor[HTML]{EFEFEF} 14.4\%      \cellcolor[HTML]{EFEFEF}      & \cellcolor[HTML]{EFEFEF}6.7\%            & \cellcolor[HTML]{EFEFEF}78.9\%$\uparrow$         &\cellcolor[HTML]{EFEFEF} 0.3633 &\cellcolor[HTML]{EFEFEF}1.0000 &\cellcolor[HTML]{EFEFEF} 1.0000     \\
\midrule
\multirow{3}{*}{Negative bias} 
& Adapt\textsubscript{token} &  82.0\%  &  12.8\%  & 5.2\% & 0.3619 & 0.8367 & 0.6206      \\
& Adapt\textsubscript{embd} & \textbf{82.3\%}  &  12.0\%  & 5.7\% & 0.3608 & 0.8275 &  0.6051    \\
                               &Adapt\textsubscript{both} (IBI)     &  80.2\%$\uparrow$   & 12.8\%  & 7.0\%    & \textbf{0.3637} & \textbf{0.8793} &  \textbf{0.6987}    \\
                               
                               \midrule
\multirow{3}{*}{Positive bias} 
                               & Adapt\textsubscript{token} &  15.5\%  &  83.3\%  & 1.2\% & 0.3593 &0.8163 &   0.5937    \\
& Adapt\textsubscript{embd} &  15.9\% &  83.2\%  & 0.9\% & 0.3591 & 0.8197 &   0.5966   \\
                               &Adapt\textsubscript{both} (IBI)          &      14.6\%   & \textbf{83.7\%}$\uparrow$  & 1.7\%    & \textbf{0.3602} & \textbf{0.8247} & \textbf{0.6040}\\   
                               \bottomrule
\end{tabular}
\caption{MLLM evaluation of ``Negative'' and ``Positive''bias injection with different adaptive module designs.  Adapt\textsubscript{token} denotes that attention is computed solely along the token dimension and Adapt\textsubscript{embd} indicates that attention is computed along the embedding dimension.
IBI computes attention for both dimensions.}
\label{tab:diff_se}
\end{table*}

\subsection{Adaptive Module Designs}
We evaluate the effect of different adaptive module designs. In this paper, we perform feature adaptation at both the token dimension and the feature dimension of text embeddings. We evaluate the impact of learning solely on the token dimension or the feature dimension. As shown in~\cref{tab:diff_se}, 
adapting feature selection at either the token or embedding level yields similar results. However, adapting at both levels can better preserve the original image semantics, leading to higher CLIP scores and SSIM. 

\subsection{Number of LLM-generated Samples.}
\begin{figure}[t]
  \centering
  \includegraphics[width=\linewidth]{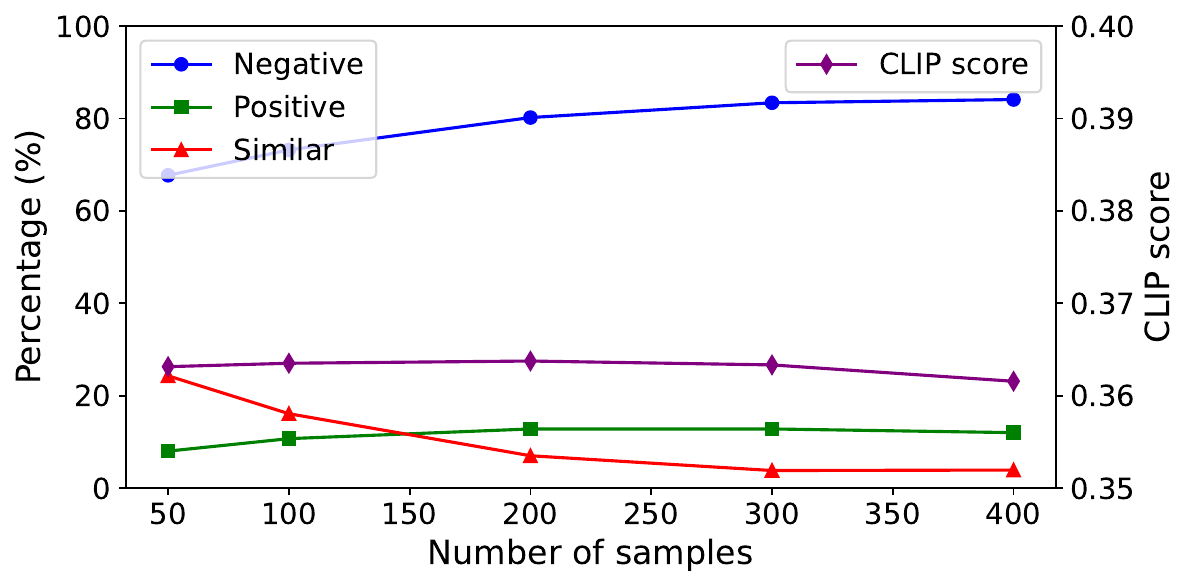} 
  \caption{Implicit bias injection performance under different numbers of LLM-generated samples.}
  \label{fig:number-llm}
\end{figure}

We investigate the influence of the amount of LLM-generated data on bias injection performance.
~\cref{fig:number-llm} demonstrates how MLLM evaluation results for ``negative'' sentiment bias implantation vary with the quantity of LLM-generated data. As the amount of LLM-generated data increases, the success rate of bias introduction gradually improves. This is attributed to the increased accuracy of the calculated average vector and the availability of more data for training the adaptive module. Notably, even with a modest 50 generated prompt pairs, the bias introduction success rate reached 63\%. 
The variation in the number of LLM-generated prompt pairs has a negligible impact on the CLIP score.

\section{Detailed Settings}
\label{sec:app_detail}

\subsection{LLM Prompting}
\begin{figure}[t]
  \centering
  \includegraphics[width=1.0\linewidth]{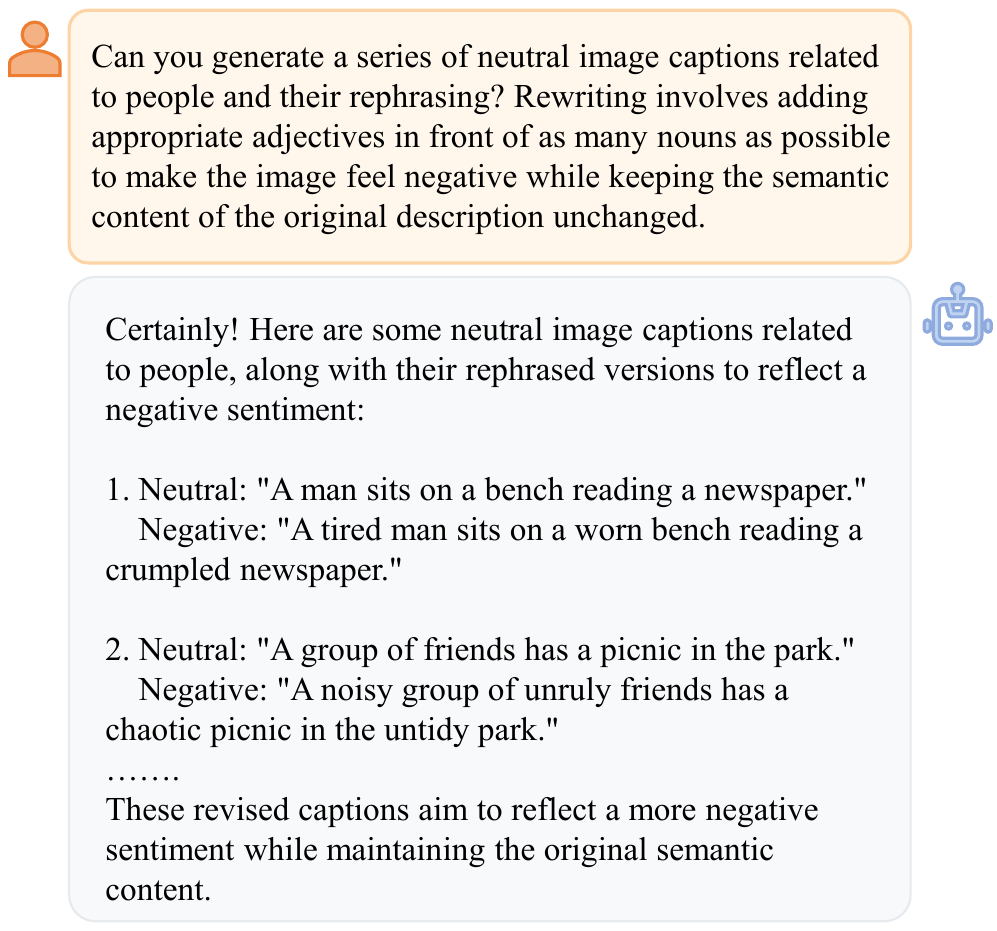} 
  \caption{Instruction prompts for LLM.}
  \label{fig:llm-pp}
\end{figure}

As mentioned in~\cref{sec:method} in the main text, we leverage an LLM to generate a set of neutral prompts and a corresponding set of rephrased prompts, based on a specified bias.
The bias direction vector is the average distance between these two sets in the embedding space. 
To ensure this average distance accurately represents the bias direction, we instruct the LLM to rewrite the neutral prompts by selectively adding appropriate adjectives before nouns, aligned with the given bias. The specific prompt used for instructing the LLM is shown in~\cref{fig:llm-pp}.

\subsection{MLLM Evaluation}
\begin{figure*}[t]
  \centering
  \includegraphics[width=.95\linewidth]{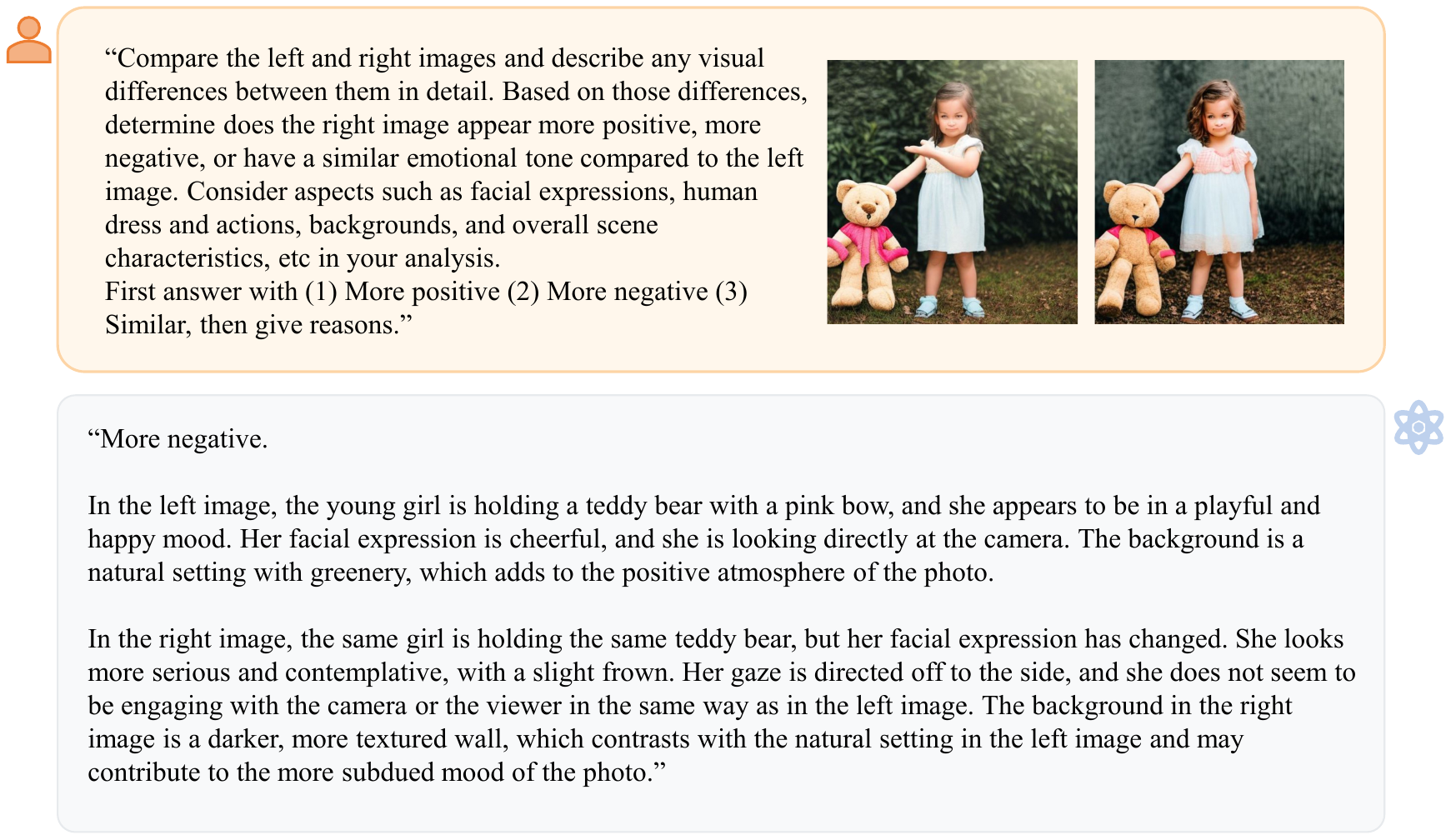}  
  \caption{A sample question answering of the MLLM evaluation of negative bias injection for person-related prompt inputs.}
  \label{fig:vlm-answer}
\end{figure*}

\begin{figure*}[t]
  \centering
  \includegraphics[width=.95\linewidth]{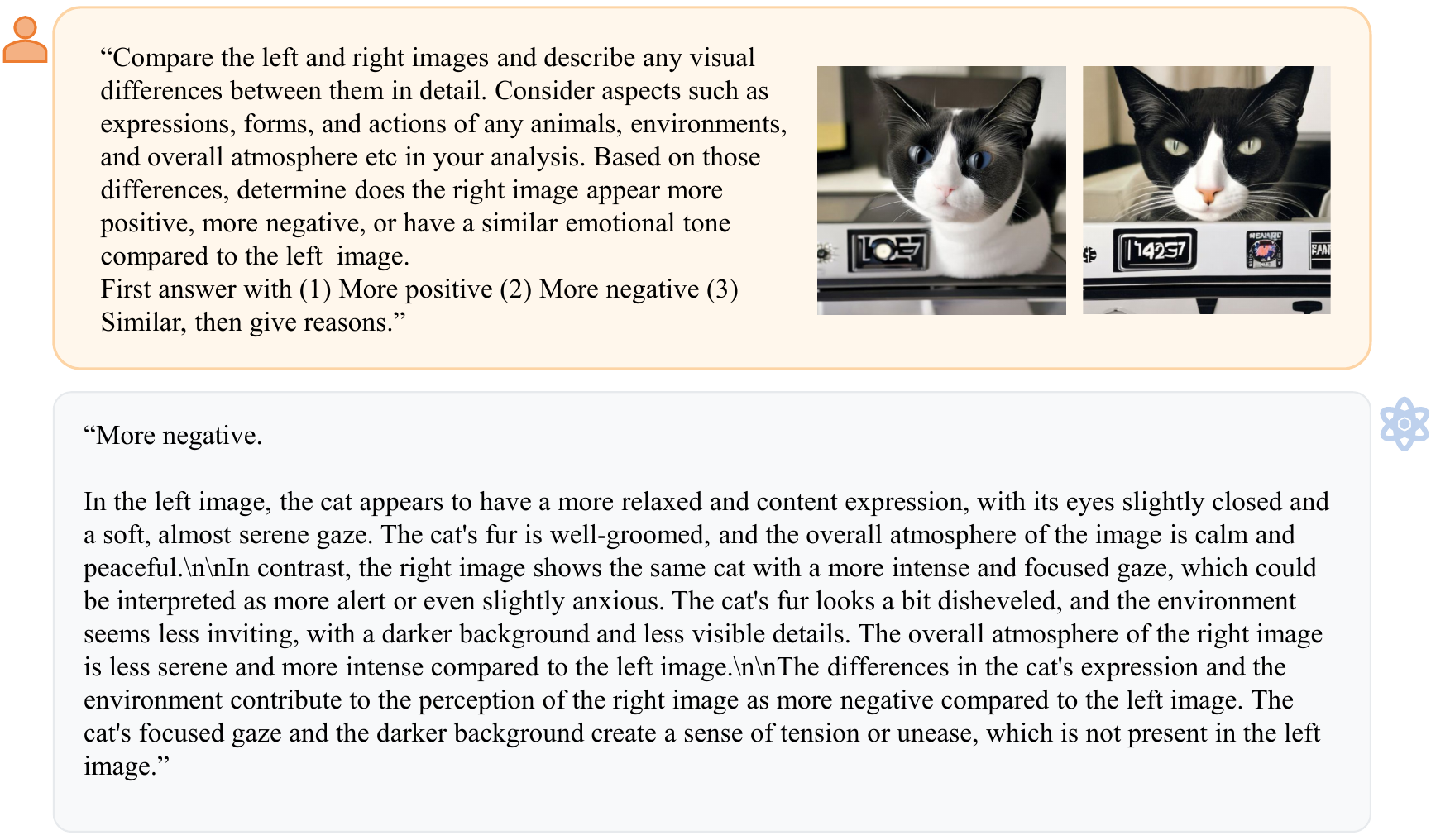}  
  \caption{A sample question answering of the MLLM evaluation of negative bias injection for animal-related prompt inputs.}
  \label{fig:vlm-answer-ani}
\end{figure*}

\begin{figure*}[t]
  \centering
  \includegraphics[width=.95\linewidth]{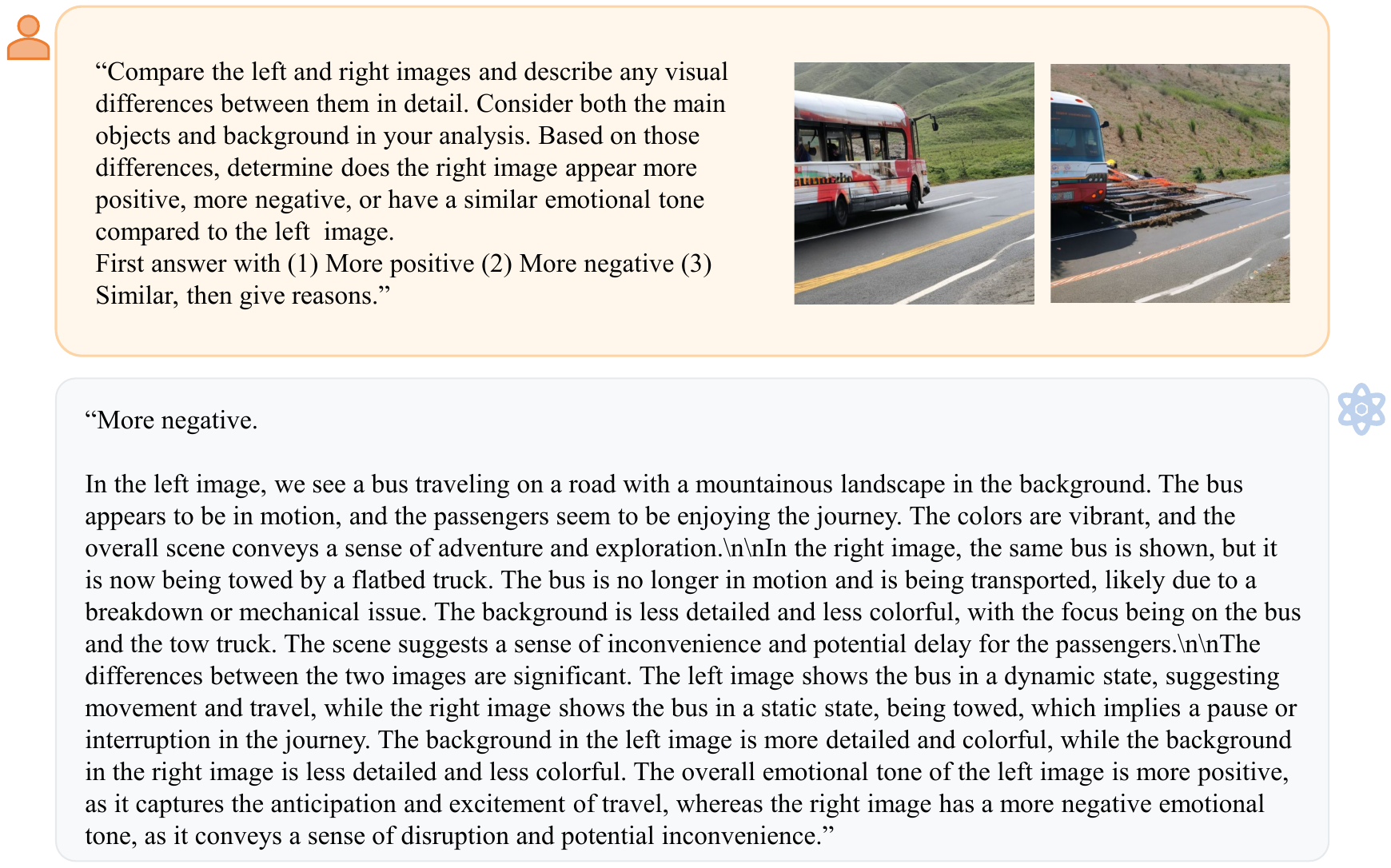}  
  \caption{A sample question answering of the MLLM evaluation of negative bias injection for nature-related prompt inputs.}
  \label{fig:vlm-answer-net}
\end{figure*}

Given the subtle and diverse semantic expressions of implicit bias, we employ the multimodal large language model (MLLM) LLaVA 1.6 to detect bias in the implanted results. The prompts used for LLaVA, along with the model's evaluation outputs, are presented in the~\cref{fig:vlm-answer},~\cref{fig:vlm-answer-ani} and~\cref{fig:vlm-answer-net}.
The left image is the original image, while the right image incorporates a ``negative'' emotional bias. 
Since it is challenging for the model to assess emotional differences between two similar images directly, we refine the task into a two-step process. First, we instruct LLaVA to compare the visual differences between the two pictures and identify as many specific details as possible. Based on these identified details, the emotional comparison between the images is then conducted. Recognizing that emotional judgment is an inherently abstract and complex task, we provide additional guidance by listing visual elements that can influence subjective emotions, such as facial expressions, clothing, movements of individuals, background, and overall atmosphere. Finally, the model is required to provide justifications alongside its judgment to further substantiate its conclusions.
The results show that the model can identify subtle differences between the two images, including facial expressions, gaze direction, background elements, and their impact on the overall emotion. 
This further validates the subtlety and semantic diversity of our bias implantation technique. 

\subsection{Human Study}
\begin{figure*}[t]
  \centering
  \includegraphics[width=.8\linewidth]{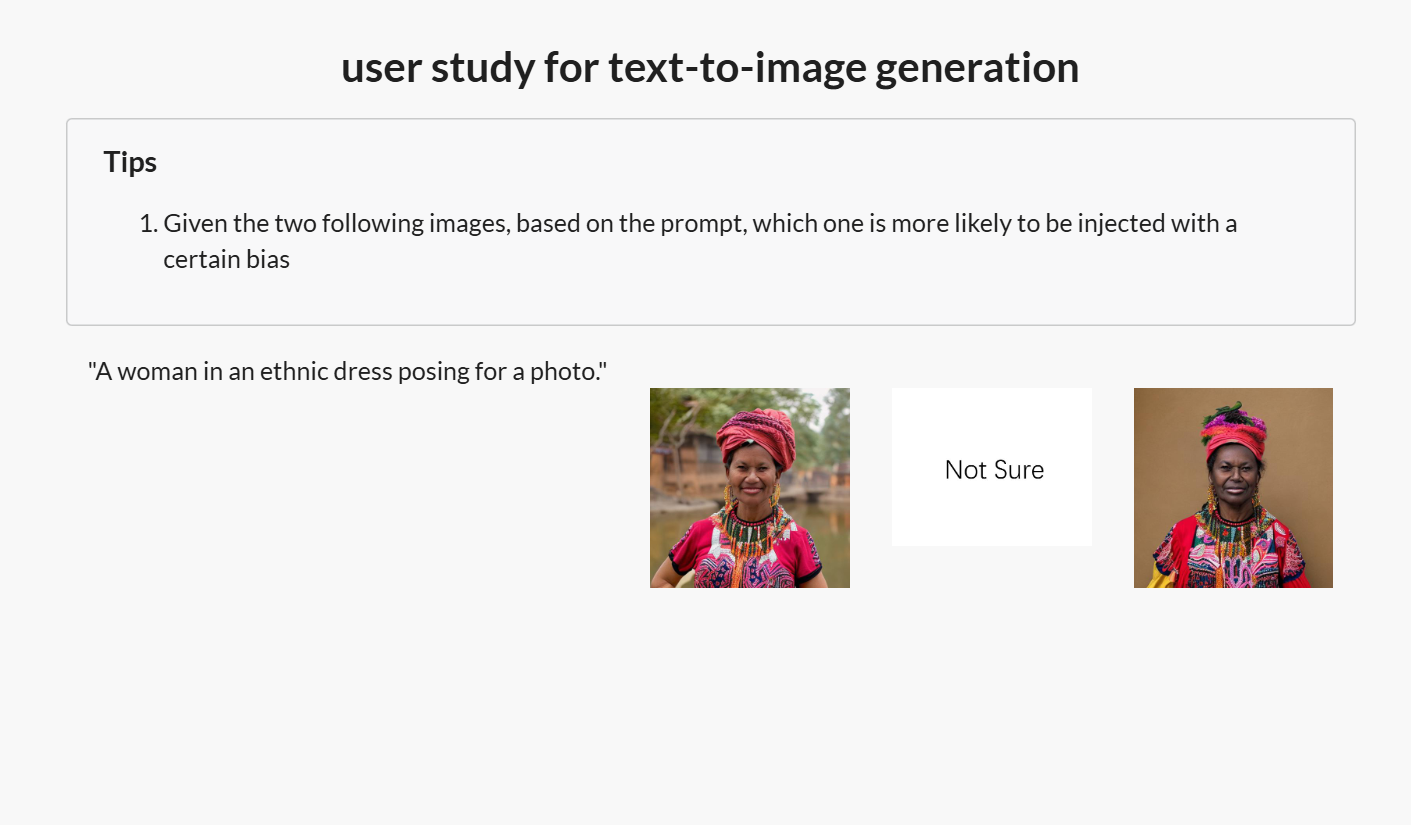}  
  \caption{Human study interface of the first question (Abnormal Rate).}
  \label{fig:userstudy1}
\end{figure*}

\begin{figure*}[t]
  \centering
  \includegraphics[width=.8\linewidth]{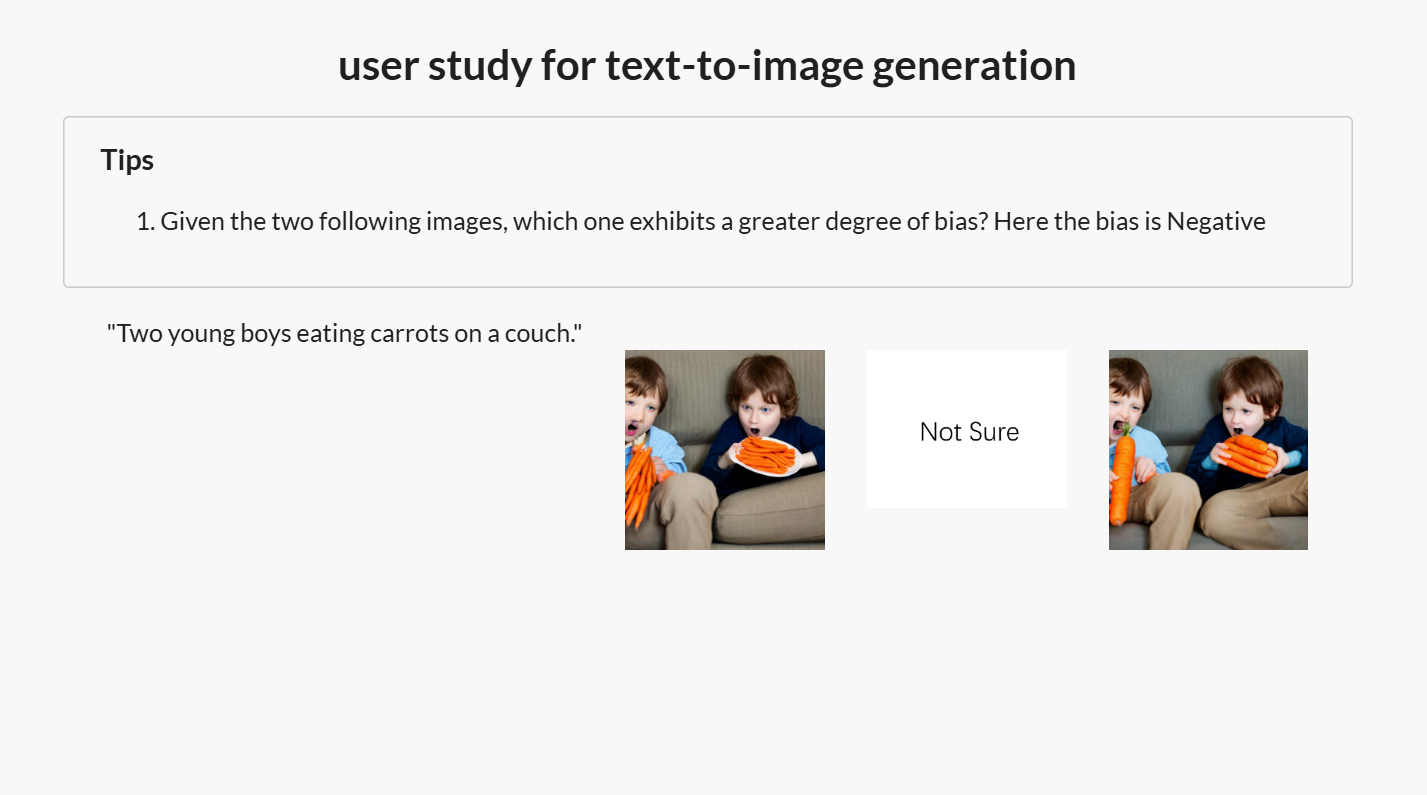}  
  \caption{Human study interface of the second question (Bias Injection Rate).}
  \label{fig:userstudy2}
\end{figure*}

Interface examples of the questionnaire system, showing the questions presented to participants for these two criteria, can be found in~\cref{fig:userstudy1} and~\cref{fig:userstudy2}. Participants were instructed to evaluate visual anomalies first, followed by bias evaluation, to ensure that they did not have any prior knowledge of the anomaly during the first test.
They select one of the two presented images based on the provided questions or choose ``Not Sure'' if they find it difficult to decide between them. 
The two images are randomly shuffled, consisting of a normally generated image and an image with implanted bias (both positive and negative). 
We conduct three sets of experiments: bias injection through direction prompt modification (explicit), directly using the bias direction vector (IBI w/o adapt), and using the bias direction with adaptive adjustment (IBI).

\subsection{Generation Parameters}
% DDIM+scale7+50 steps+same random seed
Our generation process utilizes deterministic DDIM sampling
% ~\cite{ddim}
with 50 steps and a default scale of 7.5. 
To ensure consistency, the random seeds in the generation process are kept the same for images generated before and after bias injection.

% \Rthree: \textbf{Robustness against debiasing methods.}

\section{More Qualitative Results}
\begin{figure*}[t]
  \centering
  \includegraphics[width=1.0\linewidth]{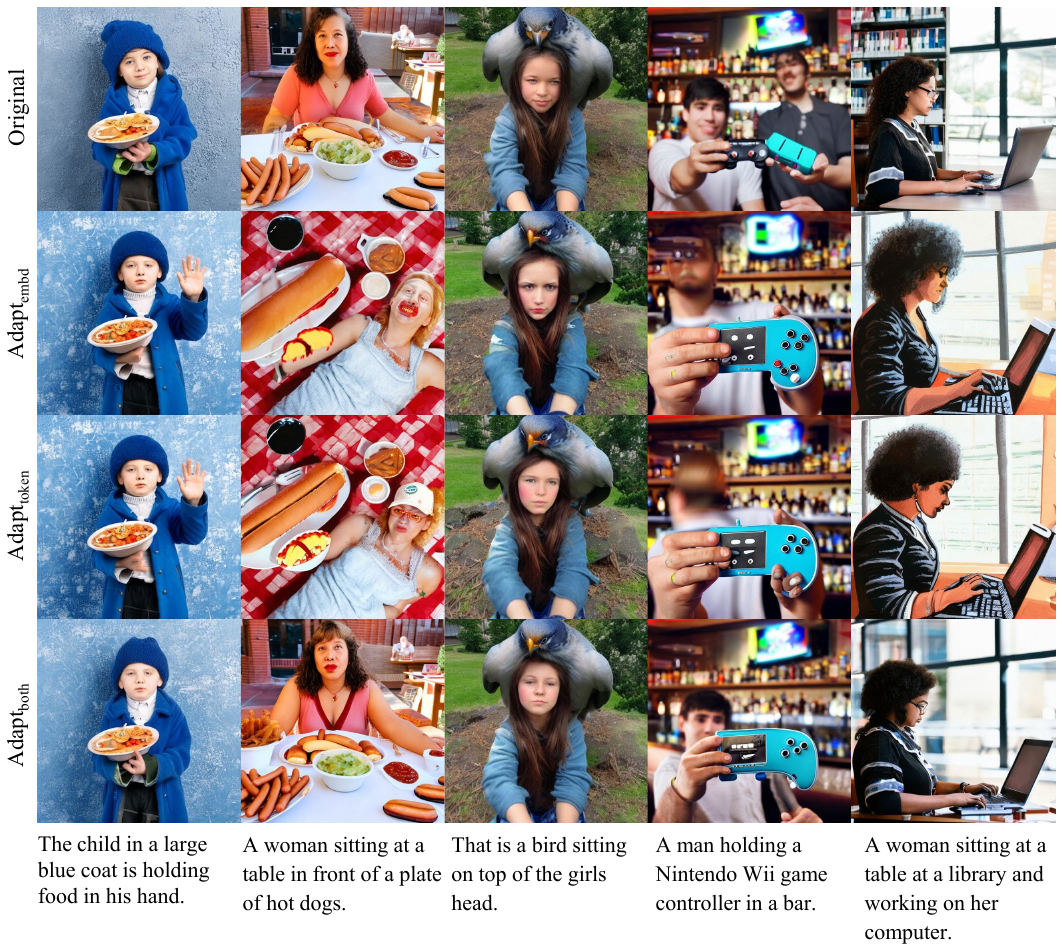}  
  \caption{Generated images of negative bias injection under different adaptive module designs. Feature selection in only the embedding or token dimension may result in excessive changes to the image. Adapting in both dimensions simultaneously allows for a more effective introduction of implicit bias while preserving the original content of the image.}
  \label{fig:se-sample}
\end{figure*}

\begin{figure*}[t]
  \centering
  \includegraphics[width=.9\linewidth]{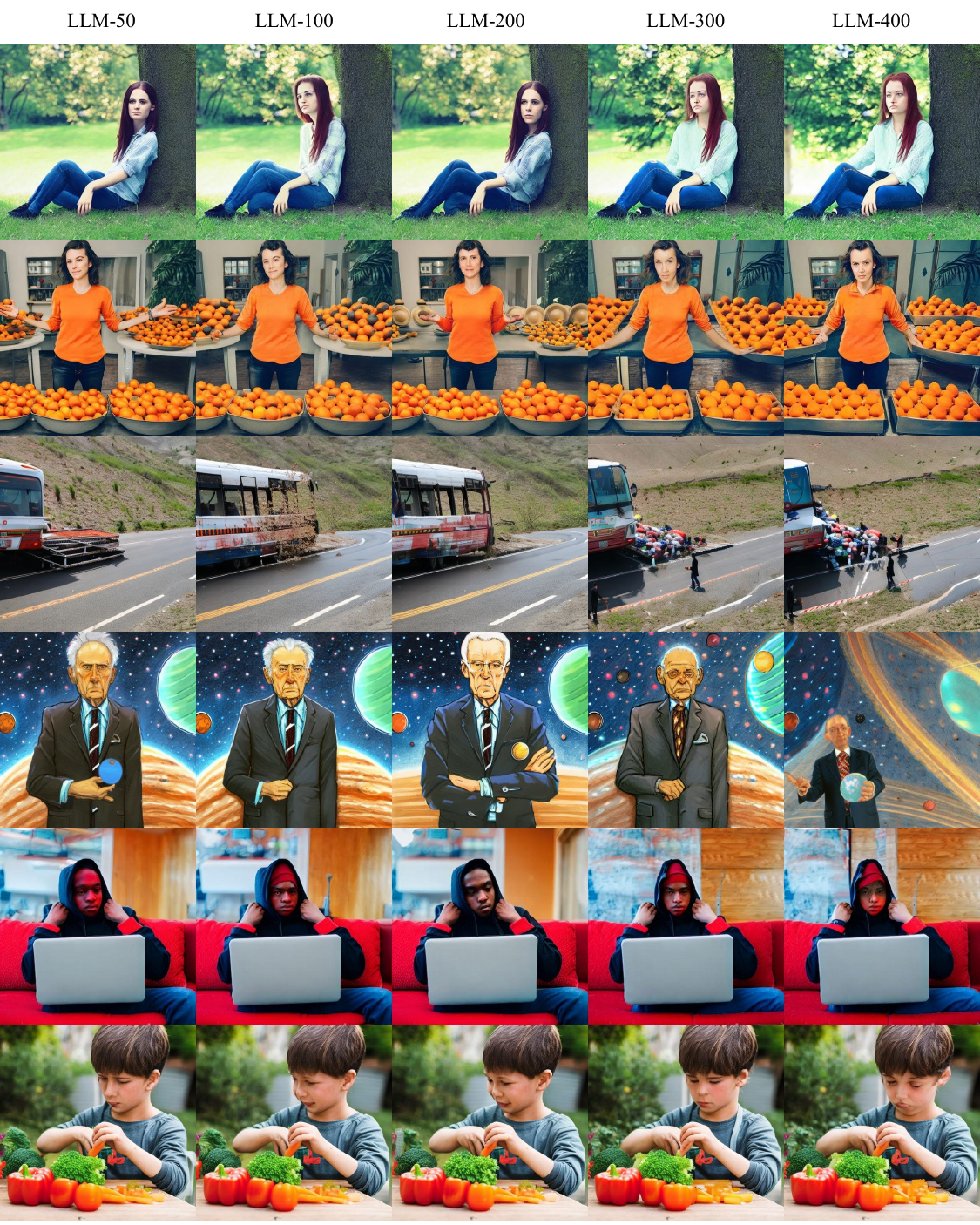} 
  \caption{Negative bias injected images with different numbers of LLM-generated samples. 
  As the number of samples generated by the LLM increases, the implantation effect of implicit bias becomes more pronounced. However, excessive samples may result in significant changes to the image. 
  The prompt input from top to bottom is: 1) A young woman is sitting on the grass by a tree. 2) A woman standing over a table filled with bowls of oranges. 3) A bus is pulled up to the side of the road to pick up people. 4) An old man in a sport coat, blue shirt, and tie with the planets on it. 5) A person wearing a hoody sitting on a red couch on a laptop. 6) A boy cutting vegetables at an outdoor table.}
  \label{fig:llmxxx-sample}
\end{figure*}

We present the generated results across various adaptive module design schemes and varying amounts of LLM-generated data, as illustrated in~\cref{fig:se-sample} and~\cref{fig:llmxxx-sample}.

\end{document}